\documentclass[journal]{IEEEtran}
\ifCLASSINFOpdf
\else
\fi
\usepackage{mathrsfs}
\usepackage{graphicx}
\usepackage{amsmath}
\usepackage{amssymb}
\usepackage{mathrsfs}
\usepackage{bm}
\usepackage{booktabs} 
\usepackage{caption}
\usepackage{subfigure}
\usepackage{algorithm}
\usepackage{algpseudocode}
\usepackage{amstext}
\usepackage{ragged2e}
\usepackage[flushleft]{threeparttable} 
\usepackage{algorithmicx}
\usepackage{algpseudocode}
\usepackage{diagbox}
\usepackage{graphics}
\usepackage{epsfig}
\usepackage{threeparttable}
\usepackage{xspace}
\usepackage{times}
\usepackage{threeparttable}
\usepackage{color}
\usepackage[normalem]{ulem}
\usepackage{multirow}
\usepackage{float}
\usepackage{amsfonts}
\usepackage{array}
\usepackage[table]{xcolor}
\usepackage{colortbl}
\usepackage{cite}
\usepackage{booktabs,multirow}

\usepackage[pagebackref=false,breaklinks=true,colorlinks,bookmarks=false]{hyperref}

\definecolor{bblue}{rgb}{0,150,230}
\definecolor{mygray}{gray}{.95}

\makeatletter
\newcommand{\thickhline}{%
    \noalign {\ifnum 0=`}\fi \hrule height 1pt
    \futurelet \reserved@a \@xhline
}
\newcommand{\midhline}{%
    \noalign {\ifnum 0=`}\fi \hrule height 0.5pt
    \futurelet \reserved@a \@xhline
}
\makeatother

\makeatletter
\DeclareRobustCommand\onedot{\futurelet\@let@token\@onedot}
\def\@onedot{\ifx\@let@token.\else.\null\fi\xspace}

\def\eg{\emph{e.g}\onedot} 
\def\ie{\emph{i.e}\onedot} 
 
\def\etc{\emph{etc}\onedot} 
 
\def\etal{\emph{et al}\onedot}

\makeatother

\usepackage{makecell}

\begin{document}
%

\title{Understanding More about Human and Machine Attention in Deep Neural Networks}
%
%
%

\author{Qiuxia Lai, Salman Khan, Yongwei Nie$^*$, Hanqiu Sun, \\Jianbing Shen,~\IEEEmembership{Senior Member,~IEEE}, Ling Shao,~\IEEEmembership{Senior Member,~IEEE}
\IEEEcompsocitemizethanks{
\IEEEcompsocthanksitem Q. Lai is with the Department of Computer Science and Engineering, The Chinese University of Hong Kong, Hong Kong, China. 
\IEEEcompsocthanksitem S. Khan is with the Mohamed bin Zayed University of Artificial Intelligence, Abu Dhabi, UAE. 
\IEEEcompsocthanksitem Y. Nie is with the School of Computer Science and Engineering, South China University of Technology, Guangzhou, China. (Email: nieyongwei@gmail.com)
\IEEEcompsocthanksitem H. Sun is with University of Electronic Science and Technology of China, Chengdu, China. (Email: sunnyqiu24@gmail.com)
\IEEEcompsocthanksitem J. Shen and L. Shao are with the Inception Institute of Artificial Intelligence, Abu Dhabi, UAE. (Email: shenjianbingcg@gmail.com)
\IEEEcompsocthanksitem $^*$Corresponding author: \textit{Yongwei Nie}
}
}

%
%

\markboth{IEEE Transactions on Multimedia}%
{Shell \MakeLowercase{\textit{et al.}}: Bare Demo of IEEEtran.cls for IEEE Journals}
%



\maketitle

\begin{abstract}

Human visual system can selectively attend to parts of a scene for quick perception, a biological mechanism known as \textit{Human attention}. Inspired by this, recent deep learning models encode attention mechanisms to focus on the most task-relevant parts of the input signal for further processing, which is called \textit{Machine/Neural/Artificial attention}. 
Understanding the relation between human and machine attention is important for interpreting and designing neural networks. 
Many works claim that the attention mechanism offers an extra dimension of interpretability by explaining where the neural networks look. However, recent studies demonstrate that artificial attention maps do not always coincide with common intuition. 
In view of these conflicting evidence, here we make a systematic study on using artificial attention and human attention in neural network design. 
With three example computer vision tasks (\ie, salient object segmentation, video action recognition, and fine-grained image classification), diverse representative backbones (\ie, AlexNet, VGGNet, ResNet) and famous architectures (\ie, Two-stream, FCN), corresponding real human gaze data, and systematically conducted large-scale quantitative studies, we quantify the consistency between artificial attention and human visual attention and offer novel insights into existing artificial attention mechanisms by giving preliminary answers to several key questions related to human and artificial attention mechanisms. 
Overall results demonstrate that human attention can benchmark the meaningful `ground-truth' in attention-driven tasks, where the more the artificial attention is close to human attention, the better the performance; for higher-level vision tasks, it is case-by-case. 
It would be advisable for attention-driven tasks to explicitly force a better alignment between artificial and human attention to boost the performance; such alignment would also improve the network explainability for higher-level computer vision tasks. 

\end{abstract}


%
\IEEEpeerreviewmaketitle

\section{Introduction}

\IEEEPARstart{H}{uman} beings can process large amounts of visual information ($10^8$-$10^9$ bits/s) in parallel through visual system~\cite{koch2006much} because the attention mechanism can selectively attends to the most informative parts of a visual stimuli rather than the whole scene~\cite{eriksen1972temporal,treisman1980feature,koch1987shifts,connor2004visual}. Different ways to mimic human visual attention have long been studied in computer vision community (dating back to~\cite{itti1998model}), since this generates more biologically inspired results, helps to understand the mechanism of human visual system~\cite{itti2001computational}, provides essential cues for downstream computer vision models~\cite{liu2017distributed,wang2018attentive,hadizadeh2012eye,gao2005discriminant,borji2015salient,sharma2012discriminative,mathe2015actions,li2017benchmark,wang2018saliency} and frees up resources to focus on the most task-related parts of inputs.

A recent trend is to integrate attention mechanisms into deep neural networks, \ie, automatically learn to selectively focus on sections of input. 
An early attempt towards an artificially attentive network was made by Tsotsos \emph{et al.}~\cite{tsotsos1995modeling} for selective tuning (ST) theory, a common attentional mechanism of spatial selection, which is further explored by STNet~\cite{biparva2017stnet} and Excitation backprop~\cite{zhang2018top}. 
Later, Bahdanau \emph{et al.}~\cite{bahdanau2014neural} incorporates attention mechnism for neural machine translation (NMT).  
Neural attention networks have shown wide success in natural language processing (NLP)~\cite{rush2015neural,wang2016attention,vaswani2017attention} and computer vision problems such as image captioning~\cite{xu2015show,cao2015look}, visual question answering (VQA)~\cite{yang2016stacked,farazi2018reciprocal}, action recognition~\cite{sharma2015action,girdhar2017attentional,zhu2018fine,du2018interaction} and salient object detection~\cite{Zhang_2018_CVPR,wang2018salient,Liu_2018_CVPR,wang2019salient,Chen_2018_ECCV,Woo_2018_ECCV}.  However, only a few works articulate precisely the relation between artificial attentions and real human attentions under certain task settings. Some efforts suggest that automatically learned attention maps can capture informative parts of the input and highlight human-sensible regions of interest~\cite{cao2015look,wang2018salient,Woo_2018_ECCV,wang2019learning}. However, recently \cite{das2016human} showed that artificial attentions do not seem to coincide with human attention for VQA task. These conflicting evidences warrant a systematic investigation into the connection between task-specific human and artificial attention.

Understanding the relation between human and machine attention is highly important as it would shed light on the reliability of artificial attention, and benchmark artificial attention against human attention, thus providing a deeper insight into the working of black-box network (or post-hoc explainability~\cite{Lipton:2018:MMI:3236386.3241340}). However, comprehensively answering the question: \textit{Whether artificial neural attention really concentrates on the meaningful parts of inputs?} is very difficult. This is mainly because: \textbf{(i)} the definition of meaningful parts is ambiguous; \textbf{(ii)} the meaningful parts are usually task-specific and can be subjective; \textbf{(iii)} it's hard to offer a quantitative evaluation due to the lack of a universally agreed `groundtruth'.

In this paper, we investigate above question by quantifying the consistency between artificial attention and human top-down attention mechanism. This is because the core motivation of artificial attention lies on human visual attention mechanism. Based on the theory of visual attention, top-down human attention is goal-driven, \ie, concentrates more on the task-relevant parts of a visual stimuli~\cite{katsuki2014bottom,itti2001computational}. Artificial attention shares similar spirit. Thus it's reasonable to explore artificial attention \textit{w.r.t} human top-down attention behavior. Additionally, in computer vision and cognitive psychology, there exist several well-established goal-driven human gaze datasets~\cite{li2014secrets,yang2013saliency,mathe2015actions,rodriguez2008action,karessli2017gaze}. During data collection, exogenous factors were controlled and the coverage of human visual attention from different subjects was guaranteed. Thus these datasets offer a relatively reliable and fairly meaningful `groundtruth' for the `informative parts' under specific task settings.
Though attention mechnism has shown promising benefits in various tasks such as monocular depth estimation~\cite{xu2018structured}, image inpainting~\cite{xie2019image}, and optical flow estimation~\cite{zhai2019optical}, here we select salient object segmentation, action recognition and fine-grained image classification as three example tasks to perform our experiments on.
That is because: \textbf{(i)} these three tasks are representative of a wide range of computer vision tasks: the first one being a pixel-wise prediction task and the remaining two being classification tasks; \textbf{(ii)} the image salient object detection is relatively \mbox{low-level and attention-driven,} while the other two are high-level vision tasks, thus covering artificial attention from different perspectives and vision levels; \textbf{(iii)} they have been accompanied with large-scale, elaborately-collected top-down human attention data~\cite{li2014secrets,yang2013saliency,mathe2015actions,rodriguez2008action,karessli2017gaze}; \textbf{(iv)} many classic network architectures (\ie, two-stream~\cite{simonyan2014two}, FCN) and backbones (\ie, AlexNet~\cite{krizhevsky2012imagenet}, VGGNet~\cite{simonyan2014very}, ResNet~\cite{he2016deep}) can be involved in our experiments, broadening an open-view towards the nature of neural attentive networks. With these tasks and gaze data, we conduct extensive quantitative and qualitative experiments with a set of attention baselines.

Our main conclusion is that there still exists a gap between neural and human attention in the three computer vision tasks. However, human attention can serve as meaningful `ground-truth' for low-level attention-driven tasks like salient object segmentation, or high-level tasks that are closely related to attention such as fine-grained image classification, where the more neural attention is close to human attention, the better the achieved performance. However, for some other higher-level vision tasks like action recognition, explicitly forcing the neural attention to mimic human attention does not bring in much improvement. 
The attention maps from different network structures and depths vary in their properties. 
Hence, we believe that an important consideration for future deep network design is to explicitly force a better alignment between artificial and human attentions for attention-driven/-related tasks to gain better performance. For other tasks, such alignment would also be a preferable way to make the decision process within deep networks more transparent and explainable.

To summarize, our contributions are four-fold:
\begin{itemize}
  \item We investigate the relation between human and machine attention by quantifying the consistency between human visual attention and artificial attention mechanisms. 
  \item We study human and neural attention mechnisms with three representative computer vision tasks, covering different attention perspectives and vision levels. 
  \item We extensively design and conduct experiments with various network backbones and architectures, followed by large-scale comparative studies. 
  \item We offer several key insights about human and neural attentions, as well as discussing future directions for designing neural attetion in deep neural networks.  
\end{itemize}

\section{Human and Artificial Attentions}\label{sec:2}
This section first provides an overview of representative works on human visual attention and famous gaze prediction datasets (\S\ref{sec:2.1}), followed by a brief overview of attention mechanisms in neural networks (\S\ref{sec:2.2}).

\vspace{-4pt}
\subsection{Human Visual Attention Mechanism}\label{sec:2.1}
\vspace{-4pt}
Human visual attention has been extensively studied for decades not only in cognitive psychology and neuroscience~\cite{past25}, but also in computer vision community (dating back to~\cite{itti1998model}). This is because such a selective visual attention mechanism has an essential role in human perception. Visual attention falls under two main categories: \textit{bottom-up} (\textit{exogenous}) and \textit{top-down} (\textit{endogenous}).

\noindent\textbf{Bottom-up Attention} is purely driven by noticeable external stimuli because of their inherent properties relative to the background~\cite{katsuki2014bottom}. 
Most early computational attention models are bottom-up methods~\cite{wang2017deep}, whose theoretical basis lies in the studies in psychology~\cite{koch1987shifts,treisman1980feature,wolfe1989guided,liu2017distributed} showing that target  stimuli ``pop out'' from their background in terms of features (\eg, color, motion, \etc) during the bottom-up attention process. More recently, several deep learning models function in a bottom-up fashion to predict a \textit{saliency map}, which is a grid-like map indicating important regions or gaze fixation distributions for the input images.

\noindent\textbf{Top-down Attention}, instead of being stimulus-inspired, is an internally induced process based on prior knowledge or goals~\cite{katsuki2014bottom,itti2001computational}. For instance, when inspecting surveillance videos, guards are more likely to allocate their attention to moving people for detecting suspicious behaviors. Endogenous attention is accompanied by longer-term cognitive factors~\cite{connor2004visual} and is very common in our daily-life~\cite{hwang2009model,pinto2013bottom,navalpakkam2006integrated}. 

\noindent\textbf{Eye-Tracking Datasets.} Most existing visual attention datasets collected gaze data during free-viewing \ie, subjects were instructed to view scenes without any particular task in mind \cite{judd2012benchmark,judd2009learning,bruce2006saliency,jiang2015salicon,mital2011clustering,Wang_2018_CVPR}. Since artificial attention is goal-directed (as detailed in \S\ref{sec:2.2}), we only consider the datasets accompanied with task-driven gaze data: \textit{PASCAL-S}~\cite{li2014secrets}, \textit{DUT-O}~\cite{yang2013saliency}, \textit{Hollywood-2}~\cite{marszalek2009actions}, \textit{UCF sports}~\cite{rodriguez2008action} and \textit{CUB-VWSW}~\cite{karessli2017gaze}. 
The former two datasets are for salient object segmentation~\cite{wang2019sodsurvey}, the following two for video action recognition, and the last for fine-grained image classification, as explained in \S\ref{sec:4}. 

\subsection{Attention Mechanism in Neural Networks}\label{sec:2.2}
Attention mechanism in neural networks, also referred to as \textit{neural/artificial attention}, can be viewed as a kind of top-down attention since it is learned in an end-to-end goal-directed manner. Such attention mechanism can be further classified as: \textit{post-hoc attention} and \textit{learned attention}.

\noindent\textbf{Post-hoc Attention.} This kind of attention maps are computed from fully trained neural networks via different strategies. Simonyan~\etal~\cite{simonyan2013deep} generated the \mbox{`class spatial attention map'} using the gradients back-propagated from the predicted score of a certain class.
Zhou~\etal~\cite{zhou2016learning} integrated global average pooling with class activation map as a proxy for attention. Post-hoc attention offers a way to assess the post-hoc explainability, \eg benefits knowledge transfer between teacher and student networks~\cite{zagoruyko2016paying}. In summary, given an existing network, these methods extract different types of attentions in a post-processing manner, to reveal the inherent reasoning process.

\noindent\textbf{Learned Attention.} Trainable neural attention can be categorized as \textit{hard} (\textit{stochastic}) and \textit{soft} (\textit{deterministic}). The former~\cite{mnih2014recurrent,xu2015show} typically needs to make hard binary choices with a low-order parameterisation. The implementation of hard attention is non-differentiable and relies on REINFORCE for its training. In this work, we concentrate on the latter class, which uses weighted average instead of hard selection and thus is fully differentiable. 
This kind of attention was first employed for ST~\cite{tsotsos1995modeling}, and was further deployed for NMT~\cite{bahdanau2014neural} in NLP and for image captioning~\cite{xu2015show} and VQA~\cite{yang2016stacked} in computer vision.
It has been shown to \mbox{perform successfully over} a wide range of computer vision tasks, such as action recognition~\cite{sharma2015action,girdhar2017attentional,song2017end,zhu2018fine,du2018interaction}, salient object detection~\cite{Zhang_2018_CVPR,wang2018salient,Liu_2018_CVPR,wang2019salient,Chen_2018_ECCV,Woo_2018_ECCV}, video object segmentation~\cite{wang2019zero} and image classification~\cite{wangresidual,jetley2018learn}.

\noindent\textbf{Relation b/w neural and human attentions.} There are few research pertaining to the connection between artificial and human attentions. Recently, Das~\etal\cite{das2016human} compared  artificial attention maps of VQA ~\cite{yang2016stacked,lu2016hierarchical} with mouse-clicked `human attention maps' collected using Amazon Mechanical Turk (AMT). Interestingly, they observed that VQA attention models do not seem to be looking at the same regions as humans. We note that their work has several limitations. 
\textit{First}, the `ground-truth' human attention maps generated by mouse-clicking may be unreliable. This is confirmed by ~\cite{bylinskii2016should}, which quantitatively showed that mouse-contingent data are noisy and do not agree well with real eye fixation data. Additionally, the data collection is performed in uncontrolled environments and under varying experimental settings (since using AMT), and the process for collation of data from different subjects is unconfirmed. Also, it is quite unclear if the collected data can accurately reflect the nature of human top-down attention considering the complex reasoning process over the high-level VQA task.
\textit{Second}, their experimental designs are ill-defined and not reliable. As an example, to exclude the center-bias effect, testing cases that have positive correlation with center attention are directly excluded, which causes significant survivorship bias. Additionally, they only use a rank-correlation measurement, which is center-bias-sensitive. However, some center-bias-resistant metrics (\ie,  shuffled AUC (s-AUC)~\cite{borji2013analysis} and Information Gain (IG)~\cite{kummerer2015information}) have already been proposed in visual attention literature.

In this work, we remedy the above limitations by performing a set of more elaborately designed experiments over three representative computer vision tasks, with more reliable human gaze data, more reasonable evaluation methodology, a complete set of baselines and an in-depth analysis. We hope that this paper, together with the work of Das~\etal\cite{das2016human}, will lead to a far richer understanding of artificial attention and motivate the research community to further explore the reliability and interpretability of artificial attention.
In our following experimental studies, we will further quantify the gap between artificial attention and human visual top-down attention, which will provide in-depth insight into the two.

\begin{figure}[!ht]
	\centering
    \includegraphics[width=\linewidth]{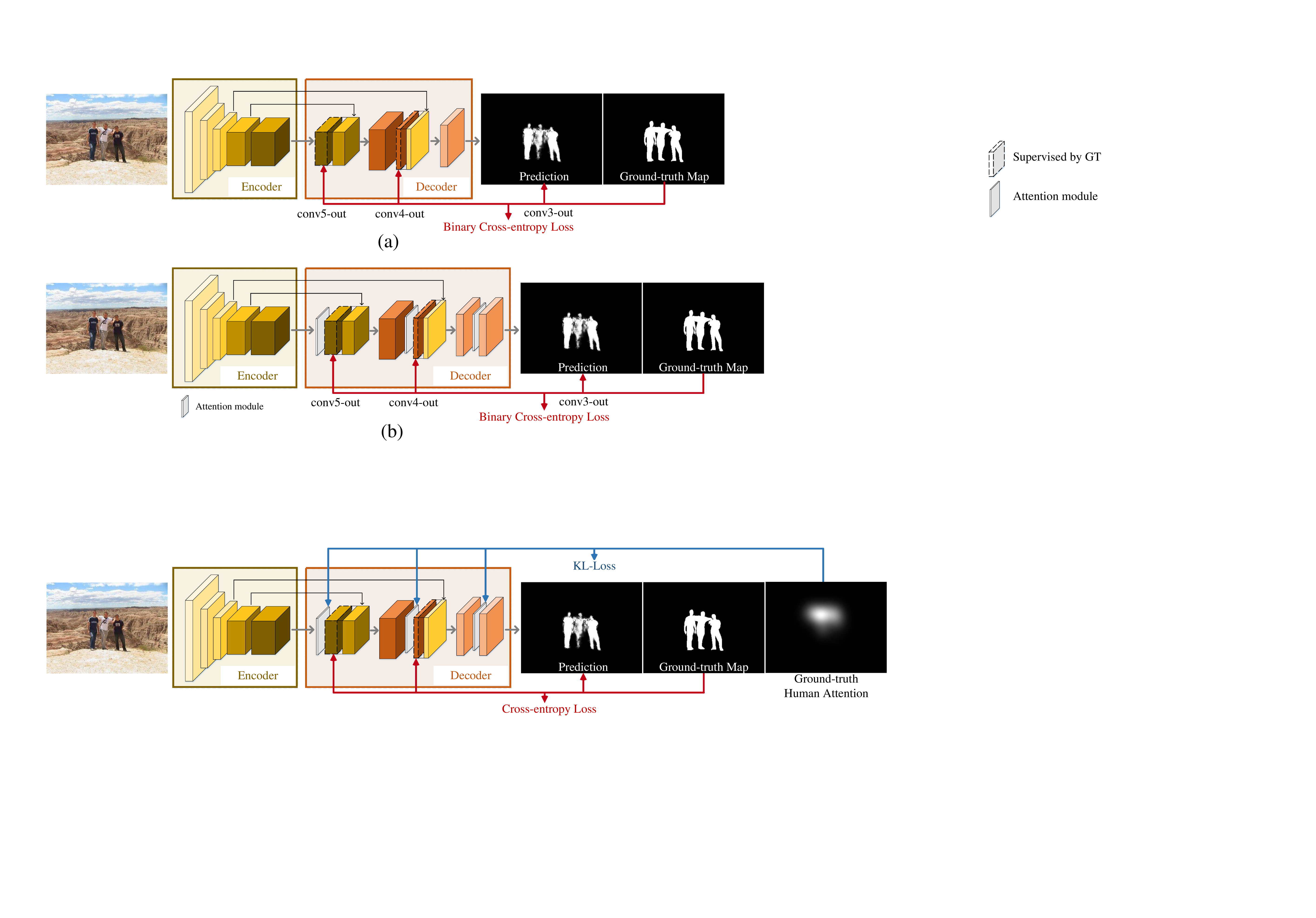}
	\caption{Illustration of network architectures for salient object detection baselines. (a) Baseline w/o. attention module; (b) Baseline w/. attention module. See \S\ref{sec:sod_network} for more details.
	\label{fig:SOD_network}}
\end{figure}

\section{Studied Neural Attention Mechanisms}\label{sec:3}

Here, we first give a general formulation of neural attention mechanism in CNN. Then, we detail three main variants, which are studied in our experiments.

Let $\mathbf{X}\!\in\!\mathbb{R}^{k\times k\times c}$ be an input tensor, $\mathbf{Z}\!\in\! \mathbb{R}^{K\times K\times D}$ a feature obtained from $\mathbf{X}$, $\mathbf{A}\!\in\! [0,1]^{K\times K}$ a soft attention map, $\mathbf{G}\!\in\! \mathbb{R}^{K\times K\times D}$ an attention glimpse and $\mathcal{F}\!: \!\mathbb{R}^{k\times k\times c}\!\rightarrow\! \mathbb{R}^{K\times K}$ an attention network that learns to map an input image to a significance matrix $\mathbf{Y} = \mathcal{F}(\mathbf{X})$. Typically, the artificial attention is implemented as:

\begin{equation}
\begin{aligned}
\mathbf{A} &= \sigma(\mathbf{Y}) = \sigma(\mathcal{F}(\mathbf{X})),\\
\mathbf{G}^d\! &= \mathbf{A}\odot\mathbf{Z}^d,
\end{aligned}
\label{eq:1}
\end{equation}
where $\sigma$ denotes an activation function that maps the significance value into $[0,1]$, and $\mathbf{G}^d$ and $\mathbf{Z}^d$ indicate the $d$-th feature slices of $\mathbf{G}$ and $\mathbf{Z}$, respectively. $\odot$ is element-wise multiplication.

\noindent\textbf{(a)~Softmax-based Neural Attention.} Here, the attention $\mathbf{A}$ is typically achieved by applying a softmax operation over all spatial locations, after learning a significance matrix $\mathbf{Y} = \mathcal{F}(\mathbf{X})$ from the input image $\mathbf{X}$:
\begin{equation}
\mathbf{A}_i = \frac{\exp(\mathbf{Y}_i)}{\sum_{j=1}^{K\times K} \exp(\mathbf{Y}_j)},
\label{eq:2}
\end{equation}
where $\mathbf{Y}\in \mathbb{R}^{K\times K}$, and $i\in 1, \ldots, K\times K$. Thus, we have $\sum_i^{K\times K}\mathbf{A}_i = 1$. This approach is called the \textit{Implicit Attention (Softmax)} in our experiments.

\noindent\textbf{(b)~Sigmoid-based Neural Attention.} Some others~\cite{Wang_2018_CVPR,Zhang_2018_ECCV} relax the sum-to-1 constraint using the \textit{sigmoid} activation function, \ie, only constrain each attention response values ranging from 0 to 1:
\begin{equation}
\mathbf{A}_i = \frac{1}{1+\exp(-\mathbf{Y}_j)}.
\label{eq:3}
\end{equation}
We name this attention mechanism as the \textit{Implicit Attention (Sigmoid)} in our experiments.

\noindent\textbf{(c)~Activation-based Post-hoc Attention.} This kind of attention was explored for network knowledge distillation~\cite{zagoruyko2016paying}. Different from the two differentiable ones above, the activation-based attention is computed during post-processing and does not contain any trainable parameters. It is constructed by computing statistics of the absolute values of the hidden feature $\mathbf{Z}$ across the channel dimension:
\begin{equation}
\mathbf{A} = \sum\nolimits_{d=1}^D|\mathbf{Z}^d|^p,
\label{eq:4}
\end{equation}
where $\mathbf{Z}^d$ indicates a slice of the feature $\mathbf{Z}$ in $d$-th channel.
We name this as \textit{Implicit Attention (Activation)}, and use it develop the comparable baseline models (\ie, \textit{w/o.} trained attention mechanisms), with $p$ set to $2$.

\begin{figure*}
\begin{minipage}[t]{0.434\textwidth}
  \centering
        \begin{threeparttable}
        \resizebox{1\textwidth}{!}{
        \setlength\tabcolsep{2.85pt}
        \renewcommand\arraystretch{1.}
        \begin{tabular}{c|c||c c |c c |c c }
        \hline\thickhline
        \rowcolor{mygray}
        &  &\multicolumn{2}{c|}{AlexNet} &\multicolumn{2}{c|}{VGGNet} & \multicolumn{2}{c}{ResNet}\\
        \cline{3-8}
        \rowcolor{mygray}
        &\multirow{-2}*{Attention} &Fmax~$\uparrow$ &MAE~$\downarrow$ &Fmax~$\uparrow$ &MAE~$\downarrow$ &Fmax~$\uparrow$ &MAE~$\downarrow$ \\
        \hline
        \hline
        \multirow{9}{*}{\rotatebox{90}{DUT-O~\cite{yang2013saliency}}}
        &Implicit attention
            &\multirow{2}*{0.604} &\multirow{2}*{0.102}  &\multirow{2}*{0.784} &\multirow{2}*{0.062} &\multirow{2}*{0.802} &\multirow{2}*{0.055}\\
        \specialrule{0em}{-0.5pt}{-2pt}
        &{\textit{(Activation)}}&&&&&&\\
        \specialrule{0em}{-0.5pt}{-1pt}
        \cline{2-8}
        &Implicit attention
            &\multirow{2}*{0.604} &\multirow{2}*{0.098} &\multirow{2}*{0.775} &\multirow{2}*{0.061} &\multirow{2}*{0.817} &\multirow{2}*{0.050}\\
        \specialrule{0em}{-0.5pt}{-2pt}
        &{\textit{(Softmax)}}&&&&&&\\
        \specialrule{0em}{-0.5pt}{-1pt}
        \cline{2-8}
        &Implicit attention
            &\multirow{2}*{0.601} &\multirow{2}*{0.099} &\multirow{2}*{0.780} &\multirow{2}*{0.059} &\multirow{2}*{0.813} &\multirow{2}*{0.050} \\
        \specialrule{0em}{-0.5pt}{-2pt}
        &{\textit{(Sigmoid)}}&&&&&&\\
        \specialrule{0em}{-0.5pt}{-1pt}
        \cline{2-8}
        &Explicit attention
            &\multirow{2}*{\textcolor{blue}{\textbf{0.607}}} &\multirow{2}*{\textcolor{blue}{\textbf{0.097}}} &\multirow{2}*{\textcolor{blue}{\textbf{0.797}}} &\multirow{2}*{\textcolor{blue}{\textbf{0.058}}} &\multirow{2}*{\textcolor{blue}{\textbf{0.820}}} &\multirow{2}*{\textcolor{blue}{\textbf{0.049}}} \\
        \specialrule{0em}{-0.5pt}{-2pt}
        &{\textit{(Supervised)}}&&&&&&\\
        \specialrule{0em}{-0.5pt}{-1pt}
        \cline{2-8}
        &Explicit attention
            &\multirow{2}*{\textcolor{red}{\textbf{0.611}}} &\multirow{2}*{\textcolor{red}{\textbf{0.095}}} &\multirow{2}*{\textcolor{red}{\textbf{0.805}}} &\multirow{2}*{\textcolor{red}{\textbf{0.056}}} &\multirow{2}*{\textcolor{red}{\textbf{0.828}}} &\multirow{2}*{\textcolor{red}{\textbf{0.047}}} \\
        \specialrule{0em}{-0.5pt}{-2pt}
        &{\textit{(Human)}}&&&&&&\\
        \specialrule{0em}{-0.5pt}{-1pt}
        \hline
        \hline
        \multirow{9}{*}{\rotatebox{90}{PASCAL-S~\cite{li2014secrets}}}
        &Implicit attention
            &\multirow{2}*{0.664} &\multirow{2}*{{0.158}} &\multirow{2}*{0.809}&\multirow{2}*{{0.096}} &\multirow{2}*{0.833} &\multirow{2}*{0.081}\\
        \specialrule{0em}{-0.5pt}{-2pt}
        &{\textit{(Activation)}}&&&&&&\\
        \specialrule{0em}{-0.5pt}{-1pt}
        \cline{2-8}
        &Implicit attention
            &\multirow{2}*{0.676} &\multirow{2}*{0.154} &\multirow{2}*{0.823}&\multirow{2}*{0.089} &\multirow{2}*{0.832 }&\multirow{2}*{0.78} \\
        \specialrule{0em}{-0.5pt}{-2pt}
        &{\textit{(Softmax)}}&&&&&&\\
        \specialrule{0em}{-0.5pt}{-1pt}
        \cline{2-8}
        &Implicit attention
            &\multirow{2}*{0.667} &\multirow{2}*{0.154} &\multirow{2}*{0.814} &\multirow{2}*{0.090} &\multirow{2}*{0.836}&\multirow{2}*{0.77} \\
        \specialrule{0em}{-0.5pt}{-2pt}
        &{\textit{(Sigmoid)}}&&&&&&\\
        \specialrule{0em}{-0.5pt}{-1pt}
        \cline{2-8}
        &Explicit attention
            &\multirow{2}*{\textcolor{blue}{\textbf{0.677}}} &\multirow{2}*{\textcolor{blue}{\textbf{0.151}}} &\multirow{2}*{\textcolor{blue}{\textbf{0.825}} }&\multirow{2}*{\textcolor{blue}{\textbf{0.088}}} &\multirow{2}*{\textcolor{blue}{\textbf{0.837}}}
            &\multirow{2}*{\textcolor{blue}{\textbf{0.076}}} \\
        \specialrule{0em}{-0.5pt}{-2pt}
        &{\textit{(Supervised)}}&&&&&&\\
        \specialrule{0em}{-0.5pt}{-1pt}
        \cline{2-8}
        &Explicit attention
            &\multirow{2}*{\textcolor{red}{\textbf{0.681}}} &\multirow{2}*{\textcolor{red}{\textbf{0.147}}} &\multirow{2}*{\textcolor{red}{\textbf{0.828}}} &\multirow{2}*{\textcolor{red}{\textbf{0.083}}} &\multirow{2}*{\textcolor{red}{\textbf{0.847}}} &\multirow{2}*{\textcolor{red}{\textbf{0.072}}} \\
        \specialrule{0em}{-0.5pt}{-2pt}
        &{\textit{(Human)}}&&&&&&\\
        \specialrule{0em}{-0.5pt}{-0pt}
        \hline
        \end{tabular}
    }
    \makeatletter\def\@captype{table}\makeatother\caption{Quantitative results of salient object detection (best in \textcolor{red}{red}, 2nd in \textcolor{blue}{blue}; same for other tables). 
    \label{table:SalObj}}
  \end{threeparttable}
  \end{minipage}
\begin{minipage}[t]{0.543\textwidth}
 \centering
    \begin{threeparttable}
        \resizebox{1\textwidth}{!}{
        \setlength\tabcolsep{1pt}
        \renewcommand\arraystretch{1.}
        \begin{tabular}{c|c||c c| c c |c c |c c| c c |c c}
        \hline\thickhline
        \rowcolor{mygray}
        &  &\multicolumn{4}{c|}{AlexNet} & \multicolumn{4}{c|}{VGGNet} & \multicolumn{4}{c}{ResNet}\\
        \cline{3-14}
        \rowcolor{mygray}
        & &\multicolumn{2}{c|}{\!s-AUC~$\uparrow$\!} &\multicolumn{2}{c|}{\!IG~$\uparrow$}
        &\multicolumn{2}{c|}{\!s-AUC~$\uparrow$\!} &\multicolumn{2}{c|}{IG~$\uparrow$}
        &\multicolumn{2}{c|}{\!s-AUC~$\uparrow$\!} &\multicolumn{2}{c}{IG~$\uparrow$} \\
        \cline{3-14}
        \rowcolor{mygray}
        &\multirow{-3}*{Attention} &best~ &worst~ &best~ &worst~ &best~ &worst~ &best~ &worst~ &best~ &worst~ &best~ &worst \\
        \hline
        \hline
        \multirow{9}{*}{\rotatebox{90}{DUT-O~\cite{yang2013saliency}}}
        &Implicit attention
            &\multirow{2}*{0.681 } &\multirow{2}*{0.572} &\multirow{2}*{0.034} &\multirow{2}*{-0.119}
            &\multirow{2}*{0.862 } &\multirow{2}*{0.578} &\multirow{2}*{0.371} &\multirow{2}*{-0.350}
            &\multirow{2}*{0.901} &\multirow{2}*{0.609}  &\multirow{2}*{0.642} &\multirow{2}*{-0.131} \\

        \specialrule{0em}{-0.5pt}{-2pt}
        &{\textit{(Activation)}}&&&&&&&&&&&\\
        \specialrule{0em}{-0.5pt}{-1pt}
        \cline{2-14}
        &Implicit attention
            &\multirow{2}*{0.738 } &\multirow{2}*{0.579} &\multirow{2}*{0.971} &\multirow{2}*{-0.664}
            &\multirow{2}*{0.787 } &\multirow{2}*{0.562} &\multirow{2}*{0.930} &\multirow{2}*{-0.167}
            &\multirow{2}*{0.725 } &\multirow{2}*{0.604}  &\multirow{2}*{0.245} &\multirow{2}*{-4.338} \\

        \specialrule{0em}{-0.5pt}{-2pt}
        &{\textit{(Softmax)}}&&&&&&&&&&&\\
        \specialrule{0em}{-0.5pt}{-1pt}
        \cline{2-14}
        &Implicit attention
            &\multirow{2}*{0.819 } &\multirow{2}*{0.607} &\multirow{2}*{1.475} &\multirow{2}*{-1.669}
            &\multirow{2}*{0.886 } &\multirow{2}*{0.644} &\multirow{2}*{1.563} &\multirow{2}*{-0.915}
            &\multirow{2}*{0.789} &\multirow{2}*{0.623}  &\multirow{2}*{0.847} &\multirow{2}*{-1.012} \\

        \specialrule{0em}{-0.5pt}{-2pt}
        &{\textit{(Sigmoid)}}&&&&&&&&&&&\\
        \specialrule{0em}{-0.5pt}{-1pt}
        \cline{2-14}
        &Explicit attention
            &\multirow{2}*{0.891 } &\multirow{2}*{0.624} &\multirow{2}*{1.825} &\multirow{2}*{-0.308}
            &\multirow{2}*{0.933 } &\multirow{2}*{0.657} &\multirow{2}*{2.207} &\multirow{2}*{-0.714}
            &\multirow{2}*{0.909 } &\multirow{2}*{0.689}  &\multirow{2}*{1.911} &\multirow{2}*{-0.365} \\

        \specialrule{0em}{-0.5pt}{-2pt}
        &{\textit{(Supervised)}}&&&&&&&&&&&\\
        \specialrule{0em}{-0.5pt}{-1pt}
        \cline{2-14}
        &Explicit attention
            &\multirow{2}*{0.940 } &\multirow{2}*{0.805} &\multirow{2}*{3.667} &\multirow{2}*{1.763}
            &\multirow{2}*{0.965 } &\multirow{2}*{0.865} &\multirow{2}*{4.165} &\multirow{2}*{2.243}
            &\multirow{2}*{0.967 } &\multirow{2}*{0.875}  &\multirow{2}*{4.238} &\multirow{2}*{2.362} \\

	\specialrule{0em}{-0.5pt}{-2pt}
	&{\textit{(Human)}}&&&&&&&&&&&\\
    \specialrule{0em}{-0.5pt}{-1pt}
        \hline
        \hline
        \multirow{9}{*}{\rotatebox{90}{PASCAL-S~\cite{li2014secrets}}} 
        &Implicit attention
            &\multirow{2}*{0.623 } &\multirow{2}*{0.578} &\multirow{2}*{0.612} &\multirow{2}*{0.402}
            &\multirow{2}*{0.722 } &\multirow{2}*{0.594} &\multirow{2}*{0.960} &\multirow{2}*{0.501}
            &\multirow{2}*{0.739} &\multirow{2}*{0.598}  &\multirow{2}*{1.026} &\multirow{2}*{0.537} \\

        \specialrule{0em}{-0.5pt}{-2pt}
        &{\textit{(Activation)}}&&&&&&&&&&&\\
        \specialrule{0em}{-0.5pt}{-1pt}
        \cline{2-14}
        &Implicit attention
            &\multirow{2}*{0.703 } &\multirow{2}*{0.609} &\multirow{2}*{1.053} &\multirow{2}*{0.691}
            &\multirow{2}*{0.740 } &\multirow{2}*{0.678} &\multirow{2}*{1.118} &\multirow{2}*{0.590}
            &\multirow{2}*{0.723} &\multirow{2}*{0.650}  &\multirow{2}*{0.535} &\multirow{2}*{-0.749} \\

        \specialrule{0em}{-0.5pt}{-2pt}
        &{\textit{(Softmax)}}&&&&&&&&&&&\\
        \specialrule{0em}{-0.5pt}{-1pt}
        \cline{2-14}
        &Implicit attention
            &\multirow{2}*{0.732 } &\multirow{2}*{0.622} &\multirow{2}*{1.605} &\multirow{2}*{0.303}
            &\multirow{2}*{0.731 } &\multirow{2}*{0.648} &\multirow{2}*{1.137} &\multirow{2}*{0.643}
            &\multirow{2}*{0.752} &\multirow{2}*{0.674}  &\multirow{2}*{1.356} &\multirow{2}*{0.131} \\

        \specialrule{0em}{-0.5pt}{-2pt}
        &{\textit{(Sigmoid)}}&&&&&&&&&&&\\
        \specialrule{0em}{-0.5pt}{-1pt}
        \cline{2-14}
        &Explicit attention
            &\multirow{2}*{0.734 } &\multirow{2}*{0.640} &\multirow{2}*{1.570} &\multirow{2}*{1.070}
            &\multirow{2}*{0.760 } &\multirow{2}*{0.663} &\multirow{2}*{1.509} &\multirow{2}*{1.276}
            &\multirow{2}*{0.760} &\multirow{2}*{0.675}  &\multirow{2}*{1.828} &\multirow{2}*{1.174} \\

        \specialrule{0em}{-0.5pt}{-2pt}
        &{\textit{(Supervised)}}&&&&&&&&&&&\\
        \specialrule{0em}{-0.5pt}{-1pt}
        \cline{2-14}
        &Explicit attention
            &\multirow{2}*{0.804 } &\multirow{2}*{0.752} &\multirow{2}*{3.047} &\multirow{2}*{2.493}
            &\multirow{2}*{0.847 } &\multirow{2}*{0.806} &\multirow{2}*{3.454} &\multirow{2}*{2.934}
            &\multirow{2}*{0.846} &\multirow{2}*{0.798}  &\multirow{2}*{3.548} &\multirow{2}*{2.782} \\

	    \specialrule{0em}{-0.5pt}{-2pt}
	    &{\textit{(Human)}}&&&&&&&&&&&\\
        \specialrule{0em}{-0.5pt}{-1pt}

        \hline
        \end{tabular}
    }
    \end{threeparttable}
    
    \makeatletter\def\@captype{table}\makeatother\caption{The correlation between attentions of top-100 and bottom-100-performance on salient object detection datasets (see \S\ref{sec:task1}).
    \label{table:SalObj_Attention_statistics}}
 \end{minipage}
\end{figure*}

\noindent\textbf{Rationale for Choice:} In this work, we only focus on the three neural attention mechanisms above. The reason is two fold. First, some variants of attention mechanisms are not suitable for our experimental settings (\eg, temporal attention). Second, other attention variants can be viewed as special cases of the above ones (\eg, channel-wise attention).
Considering our original interest in making a comprehensive comparison between artificial attention and human visual attention and yielding insights into the designing of neural attention, we intentionally consider the above typical machine attention mechanisms in our experiments (\S\ref{sec:4}).

\noindent\textbf{Relationship to Human Attention:} To offer a deeper insight into the relation between artificial attention and human visual attention, beyond the three \textit{implicit neural attentions},  we consider two \textit{explicit attention mechanisms}:

\noindent{\textbullet}~\textit{Explicit Attention} (\textit{Supervised}):~This mechanism supervises sigmoid/softmax-based neural attention with real human top-down attention. This would tell us if it is necessary to force neural attention to be close to human's for a certain task.

\noindent{\textbullet}~\textit{Explicit Attention} (\textit{Human}):~This approach directly replaces the neural attention with the real human attention, which can be viewed as the upper-bound of a human-attention-consistent neural attention model used in modern network architectures.

In the next section, we will extensively investigate neural attention by employing these five attention models with different backbones over the three example computer vision tasks.

\section{Experiments}\label{sec:4}

In this section, we experiment with the five attention baselines introduced in \S\ref{sec:3} over three example vision tasks.
To quantify the difference between artificial and human attention, we consider two metrics, shuffled AUC (s-AUC)~\cite{borji2013analysis} and Information Gain (IG)~\cite{kummerer2015information}, which are universally accepted in visual attention community, and center-bias-resistant~\cite{Bylinskiiwhat}. 

\subsection{Task1: Salient Object Segmentation}\label{sec:task1}

Salient object segmentation  aims to locate and extract the most visually important object(s) from still images. This task requires object-level understanding of the scenes. 

\subsubsection{Network Architectures}\label{sec:sod_network}

\noindent\textbf{Encoder-Decoder architecture details.}
The salient object detection model is implemented as an encoder-decoder architecture, where the encoder part is one of the three backbones which will be introduced later, and the decoder part consists of three convolutional layers for gradually making more precise pixel-wise saliency predictions, as shown in Fig.~\ref{fig:SOD_network}. The side output of each attentive feature is obtained through a \textit{Conv}($1\!\times\!1, 1$) layer with \textit{Sigmoid} activation, and supervised by the ground-truth saliency segmentation map. The final prediction comes from the $3rd$ decoder layer.
Some recent works like~\cite{Zhang_2018_CVPR,Liu_2018_CVPR} share the same essence, while appear with more complicated design.

Fig.~\ref{fig:SOD_network} (a) shows the \textit{Implicit Attention} (\textit{Activation}) baseline, where there's no attention modules incorporated in the arthitecture. For all other baselines, three attention modules are embedded layer-wisely at three decoder layers, as illustrated in Fig.~\ref{fig:SOD_network} (b). The attention module consists of a series of convolution operations, which is built as: \textit{Conv}($3\!\times\!3, \lfloor\frac{C}{2}\rfloor$) $\rightarrow$ \textit{ReLU}$\rightarrow$ \textit{Batch Normalization} (\textit{BN}) $\rightarrow$ \textit{Conv}($1\times1, 1$) $\rightarrow$ \textit{Sigmoid}, where $C$ is the channel number of the input feature. For \textit{Implicit Attention} (\textit{Softmax}) baseline, the single channel attention map is further constrained by the softmax operation over all the spatial coordinates. For \textit{Explicit Attention} (\textit{Supervised}) baseline, the three learned attention maps are supervised by the ground-truth fixation maps, while for \textit{Explicit Attention} (\textit{Human}) baseline, the attention maps are replaced directly by the top-down human visual attention.

\noindent\textbf{Backbones.}
We utilize three image classification networks as the backbones to extract the features of the input images, \ie AlexNet~\cite{krizhevsky2012imagenet}, VGG16~\cite{simonyan2014very} and ResNet50~\cite{he2016deep}.

For AlexNet, we directly use the convolutional part without any modification.
For VGGNet, we decrease the strides of the max-pooling layer in the $4th$ block to $1$, modify the dilation rates of the $5th$ convolutional block to $2$, and exclude the \textit{pool5} layer.
For ResNet, we also preserve the resolution of the final convolutional feature map by setting the strides of $4th$ and $5th$ residual blocks as $1$, and enlarging the dilation rates to be $2$ and $4$, respectively.

\subsubsection{Implementation Details}

\noindent\textbf{Datasets.}
We consider \textit{DUT-OMRON}~\cite{yang2013saliency} and \textit{PASCAL-S}~\cite{li2014secrets} in this task. With pixel-level segmentation ground-truth, these two datasets are further annotated with human gaze data.
\textit{DUT-OMRON} has $5,168$ challenging images. The fixation maps were generated from the eye-tracking data of $5$ subjects during a $2$-second viewing. Although observers are not given explicit task-related instruction during eye-tracking, but the task-irrelevant fixations are filtered out in post-processing utilizing pre-annotated bounding boxes of the salient objects, thus is resulted fixation data is implicitly affected by the high-level intention.
\textit{PASCAL-S} contains $850$ natural images with multiple objects derived from the validation set of the \textit{PASCAL-VOC 2012}~\cite{Everingham2010The}. For each image, fixations during $2$ seconds of $8$ subjects are offered. 

We perform $5$-fold cross-validation to evaluate each baseline. We randomly shuffle the image list, and divide it into five identical parts ($4$ parts for training and $1$ part for validation).
When experiment on \textit{DUT-OMRON}, we directly train on the $4$ subsets and validate on the other $1$ subset for five times. Since \textit{PASCAL-S} is relatively small, we initialized the networks with the \textit{DUT-OMRON} weights, then fine-tuned on $4$ subsets of \textit{PASCAL-S}, with the left $1$ subset as validation, where each subset is augmented $8$ times with rotation ($0^{\circ},~\!90^{\circ},~\!180^{\circ},~\!270^{\circ}$) and horizontal flipping. 

\noindent\textbf{Training details.}
The baselines of salient object detection are implemented using \textit{Keras}, and initialized with weights of ImageNet~\cite{imagenet_cvpr09}. We use \textit{Adam}~\cite{kinga2015method} to minimize the cross-entropy loss of all the saliency outputs with equal weights. For \textit{Explicit Attention} (\textit{Supervised}), the negative kl-divergence for attention maps are also minimized (weights are $0.01$). The learning rates are $10^{-5}$, $10^{-4}$ and $5\!\times\!10^{-5}$ for AlexNet, VGGNet and ResNet backbones, respectively. The inputs are scaled to $227\times227$ for AlexNet backbones, and $224\times224$ for VGGNet and ResNet backbones. The batch size is $10$.

\noindent\textbf{Evaluation Metrics.}
In salient object detection, we provide the F-measure and mean absolute error (MAE) metrics for assessing the performance of the baselines~\cite{borji2015salient}.

\textbf{\textit{F-measure.}}
F-measure comprehensively considers both precision and recall. For each image, an adaptive threshold~\cite{achanta2009frequency}, \ie twice the mean value of the saliency map, is used for generate the binary map, to calculate the precision and recall values. Then, the F-measure is calculated as a weighted harmonic mean of them, which is defined as follows:
\begin{equation}\label{equation:fbeta}
  F_{\beta}= \frac{(1+\beta^2)\text{Precision}\times \text{Recall}}{\beta^2 \text{Precision} + \text{Recall}} .
\end{equation}

\noindent We set $\beta^2$ is set to $0.3$ as suggested in~\cite{achanta2009frequency,borji2015salient}, which gives more emphasis to precision. We use the maximum $F_{\beta}$ value for measuring the performance on a dataset.

\textbf{\textit{MAE.}}
Although the above two metrics are widely used, they fail to consider the true negative pixels. The mean absolute error (MAE) is used to remedy this problem by measuring the average pixel-wise absolute error between normalized map $\mathbf{S}$ and ground-truth mask $\mathbf{G}$:
\begin{equation}\label{equation:mae}
  \text{MAE}= \frac{1}{W\!\times\!H} \sum_{w=1}^{W} \sum_{h=1}^{H} \mid \mathbf{G}(h,w)-\mathbf{S}(h,w) \mid.
\end{equation}

\noindent The mean MAE of predictions for a dataset is used to assess the performance of a salient object segmentation model.

\begin{figure*}[!ht]
	\centering
    \includegraphics[width=0.9\linewidth]{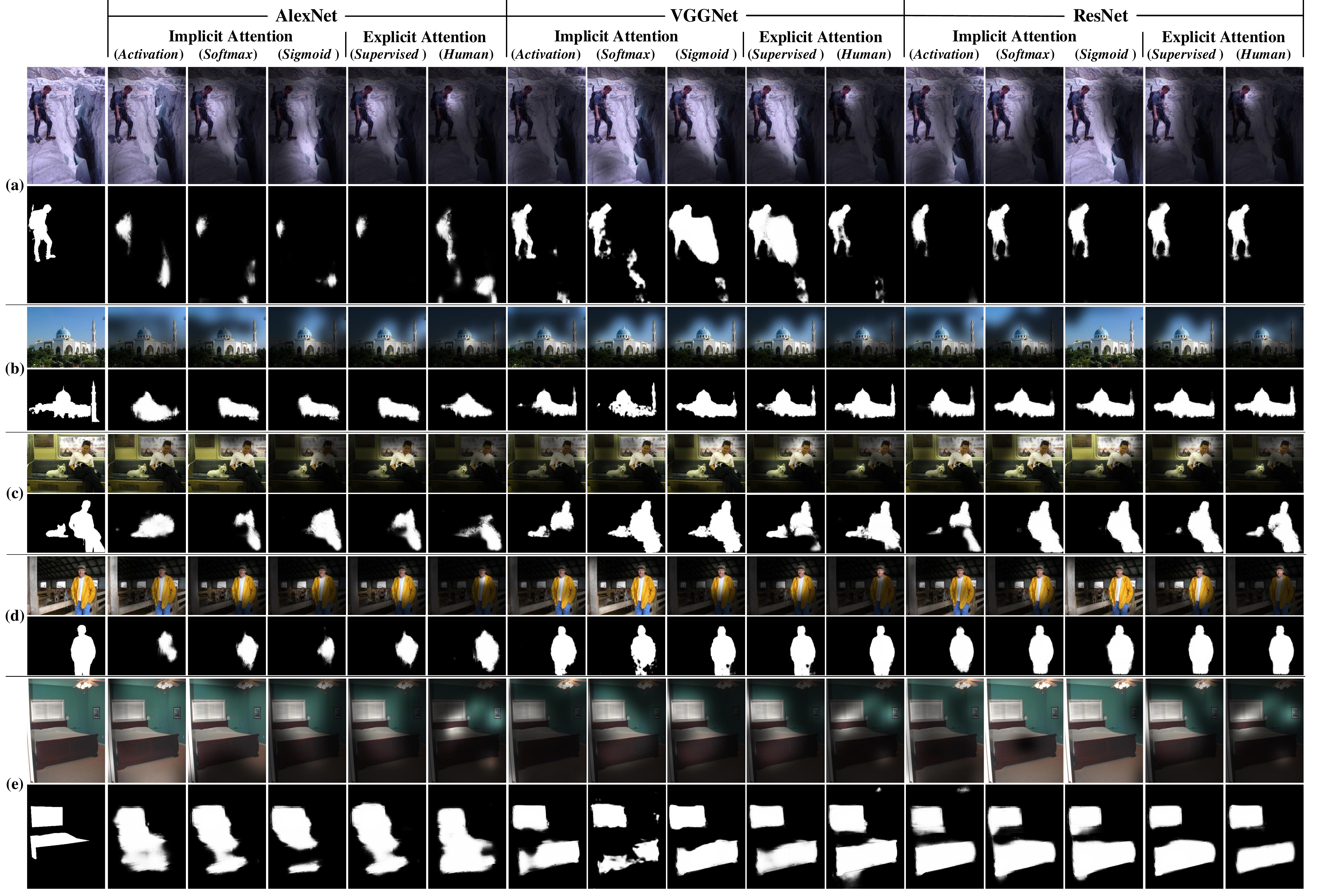}

	\caption{Qualitative results of salient object detection baselines. See \S\ref{sec:task1} for details.
	}
    \label{fig:SOD}
\end{figure*}

\subsubsection{Analyses}
\noindent\textbf{How useful is attention?} 
As shown in Table~\ref{table:SalObj}, human attention is a clear winner in this case. The explicit human attention performs better on both datasets for all three backbone architectures. When human attention is used as a supervisory signal, it helps artificial attention performs best as well. In other cases, where attention is learned implicitly, the sigmoid usually gives better performance compared to activation and softmax attentions.
Some visualized attention maps and the saliency prediction results can be found in Fig.~\ref{fig:SOD}, which shows diverse performances among various baselines with different backbones.
We can observe that:

\noindent{\textbullet}~The attention maps can help filter non-salient cues. In Fig.~\ref{fig:SOD} (a), the segmentation exclude other unrelated masses when the attention maps properly attends to the `important' part of the salient object (the meaning of `important' would be discussed in the following). For low-contrast or clustered images, the attention maps can help focus on the true `salient' part.

\noindent{\textbullet}~Complementary to the above, the attention maps can also help to include all the salient objects without missing. As shown in Fig.~\ref{fig:SOD} (b) and (c), the baselines would fail to segment the whole castle or the little dog next to the man unless the attention maps correctly highlight them.

\noindent{\textbullet}~The whole object can be inferred from a small highlighted region in the attention maps (Fig.~\ref{fig:SOD}), since the region is `important' for recognizing, \eg the face of human. 

\noindent{\textbullet}~The representation ability of the backbones have great influence on the results. First, the less effective backbone may not be able to capture human-consistent attention maps that is beneficial for saliency detection. E.g., the attention maps of AlexNet for Fig.~\ref{fig:SOD} (e) are too dispersed and do not emphasis salient part. Second, the representation ability would affect the accuracy of pixel-level predictions. As shown in Fig.~\ref{fig:SOD} (d), though AlexNet backbone generates fine attention maps, it fails to produce accurate segmentation results.

\noindent\textbf{Correlation b/w human and artificial attention.} 
We further measure how close the network attentions are to human attentions using s-AUC~\cite{borji2013analysis} and IG~\cite{kummerer2015information}. The results of the \textit{conv3} block are shown in Table~\ref{table:SalObj_Attention}. 
We conclude that the explicit human attention provides the best performance for salient object detection. The closest to direct human attention is the artificial attention supervised by human attention maps. Automatically learned artificial attention with sigmoid performs best amongst other artificial attention mechanisms, but is still significantly lower than the explicit attention mechanisms.

\noindent\textbf{Correlation b/w positive and negative cases.} We study how the attention for the best and worst performing cases correlate to each other (Table~\ref{table:SalObj_Attention_statistics}).
We draw two main conclusions. First, the correlation for explicit attention is generally stronger compared to implicit attention. Second,  the attention for best performing cases correlate better with each other compared to worst performing cases. This is because human attention itself is quite consistent and the top performing cases also become consistent in an effort to match human attention on salient object detection.

\subsection{Task2: Video Action Recognition}\label{sec:task2}
Recognizing human actions in videos is a challenging task. Here we systematically analyze how human and artificial attention can aid in action recognition. We use the recognition accuracy and the mean average precision (mAP) for single-/multi-labeled datasets, respectively.

\begin{figure*}
\begin{minipage}{\textwidth}
\begin{minipage}[t]{0.395\textwidth}
  \centering
        \begin{threeparttable}
        \resizebox{1\textwidth}{!}{
        \setlength\tabcolsep{4.3pt}
        \renewcommand\arraystretch{1.}
        \begin{tabular}{c||c|c|c|c}
        \hline\thickhline
        \rowcolor{mygray}
        & \multicolumn{2}{c|}{{Hollywood-2~\cite{marszalek2009actions}}} &\multicolumn{2}{c}{{{UCF sports~\cite{rodriguez2008action}}}}\\
        \cline{2-5}
        \rowcolor{mygray}
        &{Early Fusion} &{Late Fusion} &{Early Fusion} &{Late Fusion}\\
         \cline{2-5}
        \rowcolor{mygray}
        \multirow{-3}*{Attention}&mAP~$\uparrow$&mAP~$\uparrow$&Accuracy~$\uparrow$&Accuracy~$\uparrow$\\
        \hline
        \hline
        Implicit attention  &\multirow{2}*{0.669} &\multirow{2}*{0.675} &\multirow{2}*{0.809} &\multirow{2}*{0.681} \\
        \specialrule{0em}{-0.5pt}{-2pt}
        {\textit{(Activation)}}&\multirow{2}*{}&\multirow{2}*{}&\multirow{2}*{}&\multirow{2}*{}\\
        \specialrule{0em}{-0.5pt}{-1pt}
        \hline
        Implicit attention &\multirow{2}*{0.670} &\multirow{2}*{\textcolor{red}{\textbf{0.717}}} &\multirow{2}*{0.617} &\multirow{2}*{0.745} \\
        \specialrule{0em}{-0.5pt}{-2pt}
        {\textit{(Softmax)}}&\multirow{2}*{}&\multirow{2}*{}&\multirow{2}*{}&\multirow{2}*{}\\
        \specialrule{0em}{-0.5pt}{-1pt}
        \hline
        Implicit attention &\multirow{2}*{\textcolor{blue}{\textbf{0.707}}} &\multirow{2}*{0.676} &\multirow{2}*{\textcolor{blue}{\textbf{0.830}}} &\multirow{2}*{\textcolor{blue}{\textbf{0.787}}}\\
        \specialrule{0em}{-0.5pt}{-2pt}
        {\textit{(Sigmoid)}}&\multirow{2}*{}&\multirow{2}*{}&\multirow{2}*{}&\multirow{2}*{}\\
        \specialrule{0em}{-0.5pt}{-1pt}
        \hline
        Explicit attention &\multirow{2}*{\textcolor{blue}{\textbf{0.707}}} &\multirow{2}*{0.676} &\multirow{2}*{\textcolor{blue}{\textbf{0.830}}} &\multirow{2}*{0.617} \\
        \specialrule{0em}{-0.5pt}{-2pt}
        {\textit{(Supervised)}}&\multirow{2}*{}&\multirow{2}*{}&\multirow{2}*{}&\multirow{2}*{}\\
        \specialrule{0em}{-0.5pt}{-1pt}
        \hline
        Explicit attention &\multirow{2}*{\textcolor{red}{\textbf{0.736}}} &\multirow{2}*{\textcolor{blue}{\textbf{0.711}}} &\multirow{2}*{\textcolor{red}{\textbf{0.915}}} &\multirow{2}*{\textcolor{red}{\textbf{0.915}}} \\
        \specialrule{0em}{-0.5pt}{-2pt}
        {\textit{(Human)}}&&&&\\
        \specialrule{0em}{-0.5pt}{-0pt}
        \hline
        \end{tabular}
    }
    \end{threeparttable}
    \makeatletter\def\@captype{table}\makeatother\caption{Quantitative results of action recognition baselines. See \S\ref{sec:task2} for details.
    \label{table:ActReg}}
\end{minipage}
\begin{minipage}[t]{0.605\textwidth}
 \centering
    \begin{threeparttable}
        \resizebox{1\textwidth}{!}{
        \setlength\tabcolsep{1pt}
        \renewcommand\arraystretch{1.}
        \begin{tabular}{c||c c| c c|c c| c c |c c| c c|c c| c c}
        \hline\thickhline
        \rowcolor{mygray}
        &\multicolumn{8}{c|}{Hollywood-2~\cite{marszalek2009actions}}&\multicolumn{8}{c}{UCF sports~\cite{rodriguez2008action}}\\
        \cline{2-17}
        \rowcolor{mygray}
        &\multicolumn{4}{c|}{Early Fusion} & \multicolumn{4}{c|}{Late Fusion}&\multicolumn{4}{c|}{Early Fusion} & \multicolumn{4}{c}{Late Fusion}\\
        \cline{2-17}
        \rowcolor{mygray}
        &\multicolumn{2}{c|}{\!s-AUC~$\uparrow$\!} &\multicolumn{2}{c|}{\!IG~$\uparrow$}
        &\multicolumn{2}{c|}{\!s-AUC~$\uparrow$\!} &\multicolumn{2}{c|}{IG~$\uparrow$}
        &\multicolumn{2}{c|}{\!s-AUC~$\uparrow$\!} &\multicolumn{2}{c|}{\!IG~$\uparrow$}
        &\multicolumn{2}{c|}{\!s-AUC~$\uparrow$\!} &\multicolumn{2}{c}{IG~$\uparrow$}\\

        \cline{2-17}
        \rowcolor{mygray}
        \multirow{-4}*{Attention} &pos. &neg. &pos. &neg. &pos. &neg. &pos. &neg. &pos. &neg. &pos. &neg. &pos. &neg. &pos. &neg. \\
        \hline
        \hline
        Implicit attention
            &\multirow{2}*{0.641 } &\multirow{2}*{0.634 } &\multirow{2}*{0.247 } &\multirow{2}*{0.277 }
            &\multirow{2}*{0.716 } &\multirow{2}*{0.713 } &\multirow{2}*{0.502 } &\multirow{2}*{0.522 }
            &\multirow{2}*{0.688 } &\multirow{2}*{0.693 } &\multirow{2}*{0.879 } &\multirow{2}*{0.994 }
            &\multirow{2}*{0.721 } &\multirow{2}*{0.759} &\multirow{2}*{1.034 } &\multirow{2}*{1.324 }\\
                \specialrule{0em}{-0.5pt}{-2pt}
        {\textit{(Activation)}}&&&&&&&&&&&&&&&&\\
        \specialrule{0em}{-0.5pt}{-1pt}

        \cline{1-17}
        Implicit attention
            &\multirow{2}*{0.680 } &\multirow{2}*{0.676 } &\multirow{2}*{0.371 } &\multirow{2}*{0.408 }
            &\multirow{2}*{0.662 } &\multirow{2}*{0.667 } &\multirow{2}*{0.318 } &\multirow{2}*{0.327 }
            &\multirow{2}*{0.749 } &\multirow{2}*{0.769 } &\multirow{2}*{1.152} &\multirow{2}*{1.426}
            &\multirow{2}*{0.701 } &\multirow{2}*{0.736} &\multirow{2}*{0.811} &\multirow{2}*{1.136}\\
        \specialrule{0em}{-0.5pt}{-2pt}
        {\textit{(Softmax)}}&&&&&&&&&&&&&&&&\\
        \specialrule{0em}{-0.5pt}{-1pt}

        \cline{1-17}
        Implicit attention
            &\multirow{2}*{0.642 } &\multirow{2}*{0.643 } &\multirow{2}*{0.324 } &\multirow{2}*{0.369 }
            &\multirow{2}*{0.710 } &\multirow{2}*{0.707 } &\multirow{2}*{0.474 } &\multirow{2}*{0.493 }
            &\multirow{2}*{0.686 } &\multirow{2}*{0.671} &\multirow{2}*{1.121} &\multirow{2}*{0.952}
            &\multirow{2}*{0.720 } &\multirow{2}*{0.760} &\multirow{2}*{1.028} &\multirow{2}*{1.288}\\
        \specialrule{0em}{-0.5pt}{-2pt}
        {\textit{(Sigmoid)}}&&&&&&&&&&&&&&&&\\
        \specialrule{0em}{-0.5pt}{-1pt}

        \cline{1-17}
        Explicit attention
            &\multirow{2}*{0.677 } &\multirow{2}*{0.676 } &\multirow{2}*{0.355 } &\multirow{2}*{0.396 }
            &\multirow{2}*{0.705 } &\multirow{2}*{0.702 } &\multirow{2}*{0.457 } &\multirow{2}*{0.472 }
            &\multirow{2}*{0.751 } &\multirow{2}*{0.763 } &\multirow{2}*{1.149} &\multirow{2}*{1.423}
            &\multirow{2}*{0.712 } &\multirow{2}*{0.742 } &\multirow{2}*{1.032} &\multirow{2}*{1.629}\\
        \specialrule{0em}{-0.5pt}{-2pt}
        {\textit{(Supervised)}}&&&&&&&&&&&&&&&&\\
        \specialrule{0em}{-0.5pt}{-1pt}

        \cline{1-17}
        Explicit attention
            &\multirow{2}*{0.918 } &\multirow{2}*{0.918 } &\multirow{2}*{2.226 } &\multirow{2}*{2.424 }
            &\multirow{2}*{0.885 } &\multirow{2}*{0.881 } &\multirow{2}*{1.480 } &\multirow{2}*{1.577 }
            &\multirow{2}*{0.913 } &\multirow{2}*{0.921 } &\multirow{2}*{2.638} &\multirow{2}*{2.171}
            &\multirow{2}*{0.868 } &\multirow{2}*{0.881 } &\multirow{2}*{1.794} &\multirow{2}*{1.784}\\
        \specialrule{0em}{-0.5pt}{-2pt}
        {\textit{(Human)}}&&&&&&&&&&&&&&&&\\
        \specialrule{0em}{-0.5pt}{-0pt}
        \hline
        \end{tabular}
    }
    \end{threeparttable}
    \makeatletter\def\@captype{table}\makeatother\caption{The correlation between positive and negative attention maps on action recognition datasets (see \S\ref{sec:task2}). 
    \label{table:ActRec_Attention_statistics}}
 \end{minipage}
 \end{minipage}
\end{figure*}

\subsubsection{Network Architectures}\label{sec:ar_network}
We use two-stream~\cite{simonyan2014two} and 3D ConvNet~\cite{tran2015learning} architecture for building up the action recognition architecture. We further study two typical attention embedding strategies: early fusion (embedding attention between the first convolution blocks of the frame stream and the optical flow stream) and late fusion (the attention module is embedded late with the fusion of the last convolution features).

\noindent\textbf{3D ConvNet.}
C3D network~\cite{tran2015learning} is consisted of 3D building blocks, such as 3D convolution and 3D pooling, which explicitly operate along the time dimension for processing the motion information. 
We apply the same structure for the spatial and temporal streams in the two-stream architecture.

\begin{table}[t]
    \centering
    \begin{threeparttable}
        \resizebox{0.48\textwidth}{!}{
        \setlength\tabcolsep{3pt}
        \renewcommand\arraystretch{1.}
        \begin{tabular}{c|c||c c| c c |c c}
        \hline\thickhline
        \rowcolor{mygray}
        &  &\multicolumn{2}{c|}{AlexNet} & \multicolumn{2}{c|}{VGGNet} & \multicolumn{2}{c}{ResNet}\\
        \cline{3-8}
        \rowcolor{mygray}
        &\multirow{-2}*{Attention}  &s-AUC~$\uparrow$ &IG~$\uparrow$ &s-AUC~$\uparrow$ &IG~$\uparrow$ &s-AUC~$\uparrow$ &IG~$\uparrow$ \\
        \hline
        \hline
         \multirow{9}{*}{\rotatebox{90}{DUT-O~\cite{yang2013saliency}}}
        &Implicit attention
            &\multirow{2}*{0.636 }&\multirow{2}*{-0.096} &\multirow{2}*{0.707} &\multirow{2}*{0.138} &\multirow{2}*{0.726} &\multirow{2}*{0.042} \\
        \specialrule{0em}{-0.5pt}{-2pt}
        &{\textit{(Activation)}}&&&&&&\\
        \specialrule{0em}{-0.5pt}{-1pt}
        \cline{2-8}
        &Implicit attention
            &\multirow{2}*{0.674} &\multirow{2}*{0.284} &\multirow{2}*{0.705} &\multirow{2}*{0.319} &\multirow{2}*{0.699} &\multirow{2}*{0.220}\\
        \specialrule{0em}{-0.5pt}{-2pt}
        &{\textit{(Softmax)}}&&&&&&\\
        \specialrule{0em}{-0.5pt}{-1pt}
        \cline{2-8}
        &Implicit attention
            &\multirow{2}*{0.720} &\multirow{2}*{0.521} &\multirow{2}*{0.761} &\multirow{2}*{0.555} &\multirow{2}*{0.704}&\multirow{2}*{0.165} \\
        \specialrule{0em}{-0.5pt}{-2pt}
        &{\textit{(Sigmoid)}}&&&&&&\\
        \specialrule{0em}{-0.5pt}{-1pt}
        \cline{2-8}
        &Explicit attention
            &\multirow{2}*{\textcolor{blue}{\textbf{0.748}}} &\multirow{2}*{\textcolor{blue}{\textbf{0.986}}} &\multirow{2}*{\textcolor{blue}{\textbf{0.789}}} &\multirow{2}*{\textcolor{blue}{\textbf{1.116}}} &\multirow{2}*{\textcolor{blue}{\textbf{0.791}}} &\multirow{2}*{\textcolor{blue}{\textbf{1.101}}} \\
        \specialrule{0em}{-0.5pt}{-2pt}
        &{\textit{(Supervised)}}&&&&&&\\
        \specialrule{0em}{-0.5pt}{-1pt}
        \cline{2-8}
        &Explicit attention
            &\multirow{2}*{\textcolor{red}{\textbf{0.861}}} &\multirow{2}*{\textcolor{red}{\textbf{2.465}}} &\multirow{2}*{\textcolor{red}{\textbf{0.901}}} &\multirow{2}*{\textcolor{red}{\textbf{2.839}}} &\multirow{2}*{\textcolor{red}{\textbf{0.900}}} &\multirow{2}*{\textcolor{red}{\textbf{2.839}}} \\
  \specialrule{0em}{-0.5pt}{-2pt}
  &{\textit{(Human)}}&&&&&&\\
\specialrule{0em}{-0.5pt}{-1pt}
        \hline
        \hline
        \multirow{9}{*}{\rotatebox{90}{PASCAL-S~\cite{li2014secrets}}}
        &Implicit attention
            &\multirow{2}*{0.590} &\multirow{2}*{0.504} &\multirow{2}*{0.640} &\multirow{2}*{0.627} &\multirow{2}*{0.650} &\multirow{2}*{0.722 }\\
        \specialrule{0em}{-0.5pt}{-2pt}
        &{\textit{(Activation)}}&&&&&&\\
        \specialrule{0em}{-0.5pt}{-1pt}
        \cline{2-8}
        &Implicit attention
            &\multirow{2}*{0.636} &\multirow{2}*{0.746} &\multirow{2}*{0.689} &\multirow{2}*{0.789} &\multirow{2}*{0.681} &\multirow{2}*{0.138} \\
        \specialrule{0em}{-0.5pt}{-2pt}
        &{\textit{(Softmax)}}&&&&&&\\
        \specialrule{0em}{-0.5pt}{-1pt}
        \cline{2-8}
        &Implicit attention
            &\multirow{2}*{0.658} &\multirow{2}*{0.746} &\multirow{2}*{0.676} &\multirow{2}*{0.787} &\multirow{2}*{0.698} &\multirow{2}*{0.794} \\
        \specialrule{0em}{-0.5pt}{-2pt}
        &{\textit{(Sigmoid)}}&&&&&&\\
        \specialrule{0em}{-0.5pt}{-1pt}
        \cline{2-8}
        &Explicit attention
            &\multirow{2}*{\textcolor{blue}{\textbf{0.678}}} &\multirow{2}*{\textcolor{blue}{\textbf{1.343}}} &\multirow{2}*{\textcolor{blue}{\textbf{0.698}}} &\multirow{2}*{\textcolor{blue}{\textbf{1.357}}} &\multirow{2}*{\textcolor{blue}{\textbf{0.707}}}&\multirow{2}*{\textcolor{blue}{\textbf{1.406}}} \\
        \specialrule{0em}{-0.5pt}{-2pt}
        &{\textit{(Supervised)}}&&&&&&\\
        \specialrule{0em}{-0.5pt}{-1pt}
        \cline{2-8}
        &Explicit attention
             &\multirow{2}*{\textcolor{red}{\textbf{0.773}}} &\multirow{2}*{\textcolor{red}{\textbf{2.656}}} &\multirow{2}*{\textcolor{red}{\textbf{0.816}}} &\multirow{2}*{\textcolor{red}{\textbf{3.058}}} &\multirow{2}*{\textcolor{red}{\textbf{0.817}}} &\multirow{2}*{\textcolor{red}{\textbf{3.058}}} \\
    \specialrule{0em}{-0.5pt}{-2pt}
    &{\textit{(Human)}}&&&&&&\\
            \specialrule{0em}{-0.5pt}{-0pt}
        \hline
        \end{tabular}
    }
    \end{threeparttable}
    \caption{The correlation between human and neural attentions on salient object segmentation datasets (see \S\ref{sec:task1}).
    \label{table:SalObj_Attention}}
\end{table}

\noindent\textbf{Early attention in two-stream architecture.}
Early attention module acts before two-stream fusion. We study the $5$ implicit and explicit attention baselines. The attention is applied at the first convolution block of the frame stream, either generated from the corresponding block of the optical flow stream, or replaced by the ground truth human visual attention.

\noindent\textbf{Late attention in two-stream architecture.}
For late fusion, we implement the \textit{ReLU5+FC8} architecture where both convolutional fusions and fully-connected fusions are performed to guarantee the learning ability of the two-stream structure. The attention is applied at the fusion feature map of the \textit{relu\_5b}, where both spatial and temporal resolution has been reduced after 3D pooling.

\begin{figure*}
    \centering
    \includegraphics[width=0.7\linewidth]{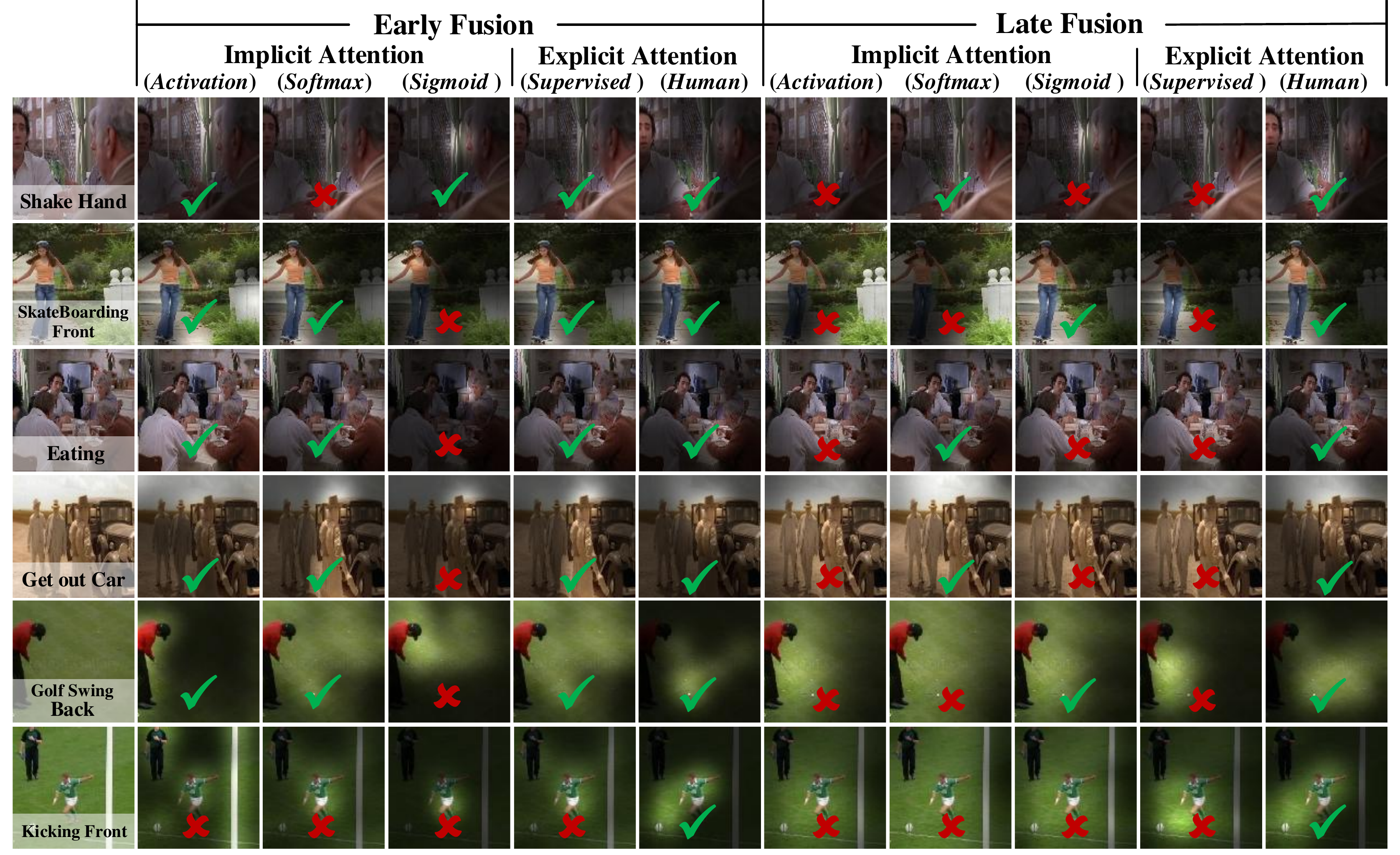}

    \caption{Qualitative results of action recognition baselines (see \S\ref{sec:task2}).  
    \label{fig:AR}}
\end{figure*}

\subsubsection{Implementation Details}

\noindent\textbf{Datasets.}
We conduct experiments on \textit{Hollywood-2}~\cite{marszalek2009actions} and \textit{UCF sports}~\cite{rodriguez2008action}.  \textit{Hollywood-2} comprises $1,707$ video sequences, which are collected from $69$ Hollywood movies from $12$ action categories, such as eating, kissing and running.  \textit{UCF sports} contains $150$ videos, which cover $9$ common sports action classes, such as diving, swinging and walking. Mathe~\etal~\cite{mathe2015actions} annotated these two datasets with task-driven gaze data. The fixation data were collected from $19$ observers belonging to $3$ groups for free viewing ($3$ observers), action recognition ($12$ observers), and context recognition ($4$ observers). 

To get training data for the two-stream architecture, we extract the frames and compensated optical flows images~\cite{zach2007duality}, and build the file list of video snippets\footnote{Code in curtesy of https://github.com/antran89/two-stream-fcan}. We set split number to $1$, and length of consecutive frames/optical flows to be $16$.

\noindent\textbf{Training details.}
We implement all the baselines using caffe~\cite{jia2014caffe}. For implicit baselines, we optimize the cross-entropy error function. For explicit baseline, we incorporate an extra softmax multinomial logistic loss. 
We choose mini-batch stochastic gradient descent (SGD)~\cite{lecun1998gradient} as the solver.

To avoid overfitting on \textit{UCF sports}, which is relatively small, we first train the model on \textit{Hollywood-2}, then fine-tuned on \textit{UCF sports}.
Considering the complex network architecture, we trained the models in tandem on \textit{Hollywood-2}. The spatial stream is directly initialized with the weight trained on ImageNet~\cite{imagenet_cvpr09}. For the temporal stream, we initialized with C3D Sports1M~\cite{tran2015learning} weights and further fine-tuned on the video compensated flow data for $1K$ iterations with initial learning rate being $5\times 10^{-3}$ which is multiplied by $0.1$ at $4K$ and $6K$ iterations.
When training the two-stream architecture, the batch size is $32$.
The initial learning rate is set to be $10^{-4}$, and is twice decreased with a factor of $0.1$ at the $4K$ and $8K$ iterations, respectively, out of $10K$ iterations. When fine-tuning on \textit{UCF sports}, the initial learning rate is set to be $10^{-6}$ (early fusion) or $10^{-5}$ (late fusion), and multiplied by $0.1$ at the $10K$ and $20K$ iterations out of $30K$ iterations.

\noindent\textbf{Evaluation Metrics.}
In video action recognition, we evaluate the mAP for \textit{Hollywood-2} (multi-label classification), and the accuracy for \textit{UCF sports} (single-label classification).

\textbf{\textit{mAP.}}
First, let us recall the mathematical definition of precision $p$ and recall $r$:
\begin{equation}\label{equation:pr}
   \begin{aligned}
        p &= \frac{\# true\;positive}{\# true\;positive + \# false\;positive}, \\
        r &= \frac{\# true\;positive}{\# true\;positive + \# false\;negative}.
   \end{aligned}
\end{equation}

The mAP is calculated as the mean of the averaged precision (AP) of all the classes. For each class, AP is calculated as the mean of maximum precision at each recall level. To be concrete, when plotting the recall-precision curve, the precision value $p_{interp}(r)$ at recall $r$ is replaced with the maximum precision for any recall $\tilde{r}>r$:
\begin{equation}\label{equation:pinterp}
  p_{interp}(r) = \max\limits_{\tilde{r}\geq r}p(\tilde{r})  .
\end{equation}
For action recognition, AP for a certain class can be calculated as the average of the interpolated precisions at all the recall levels (\textit{i.e.} all the videos predicted as this class).
Then, the mAP is the mean of APs for all classes.

\textbf{\textit{Accuracy.}}
Here, we calculate the accuracy based on video. The prediction of a certain video is obtained by averaging the predictions of all the tested frames in that video and get the class index with maximum probability.
\begin{equation}\label{equation:acc}
  Accuracy= \frac{\# correct\;classified\;videos}{\# total\;testing\;videos}.
\end{equation}
The $\# total\;testing\;videos$ is $47$ for \textit{UCF sports}.

\subsubsection{Analyses}
\noindent\textbf{How useful is attention?} 
From the quantitative results in Table~\ref{table:ActReg}, we notice four key observations: \textbf{(1)} The explicit attention from human generally outperforms the artificial attention in neural networks. \textbf{(2)} The use of explicit attention for supervised training does not add much value compared to implicitly learned attention mechanisms (esp. softmax and sigmoid variants). \textbf{(3)} Among the implicit artificial attention mechanisms, sigmoid-based neural attention generally outperforms other variants, which is consistent with Task1 (\S\ref{sec:task1}). \textbf{(4)} Activation based implicit attention performs lowest. This trend is understandable because activation based attention is computed in a post-hoc manner and does not benefit from training process.

Fig.~\ref{fig:AR} shows the visualization for different attention maps on several examples.
From the visualization of the attention maps, we may conclude that the consistency between the artificial and human attentions is not the key factors for the correctness of action recognition. It is complex to determine whether the neural attentions look at `meaningful' parts of the image in high-level vision tasks like action recognition, since human attention does not have decisive influence on the performance. However, explicitly forcing the alignment between artificial and human attentions would not decrease the performance generally; on the other hand, it would make the deep networks more transparent and explainable.

\noindent\textbf{Correlation b/w human and artificial attention.} We study the correlation between artificial and human attention using s-AUC~\cite{borji2013analysis} and IG~\cite{kummerer2015information} (Table~\ref{table:ActRec_Attention}). 
Among neural attention maps, explicit attention trained with supervised human attention yields the most consistent performance across the two metrics and for different fusion mechanisms (early and late), on both datasets. 
This trend arises because the network directly learns to predict attention maps close to human attention. Among implicit attention mechanisms, surprisingly, the activation-based attention correlates quite well with human attention. This shows that even the post-hoc attention maps relate well with human attention \ie, the network indirectly learns to focus on important details in a scene.

\begin{table}[t]
    \centering
         \centering
\resizebox{0.48\textwidth}{!}{
        \setlength\tabcolsep{1pt}
        \renewcommand\arraystretch{1.}
        \begin{tabular}{c||c c| c c|c c| c c}
        \hline\thickhline
        \rowcolor{mygray}
        &\multicolumn{4}{c|}{Hollywood-2~\cite{marszalek2009actions}}&\multicolumn{4}{c}{UCF sports~\cite{rodriguez2008action}}\\
        \cline{2-9}
        \rowcolor{mygray}
        &\multicolumn{2}{c|}{Early Fusion} & \multicolumn{2}{c|}{Late Fusion}&\multicolumn{2}{c|}{Early Fusion} & \multicolumn{2}{c}{Late Fusion}\\
        \cline{2-9}
        \rowcolor{mygray}
        \multirow{-3}*{Attention}  &s-AUC~$\uparrow$\! &IG~$\uparrow$ &s-AUC~$\uparrow$\! &IG~$\uparrow$ &s-AUC~$\uparrow$\! &IG~$\uparrow$ &s-AUC~$\uparrow$\! &IG~$\uparrow$\\
        \hline
        \hline
         Implicit attention
            &\multirow{2}*{0.640 } &\multirow{2}*{0.249} &\multirow{2}*{\textcolor{blue}{\textbf{0.713}} } &\multirow{2}*{\textcolor{blue}{\textbf{0.503}} }
            &\multirow{2}*{0.687 } &\multirow{2}*{0.872 } &\multirow{2}*{\textcolor{blue}{\textbf{0.730}} } &\multirow{2}*{1.095 }\\
                \specialrule{0em}{-0.5pt}{-2pt}
        {\textit{(Activation)}}&&&&&&&&\\
        \specialrule{0em}{-0.5pt}{-1pt}
        \cline{1-9}
        Implicit attention
            &\multirow{2}*{\textcolor{blue}{\textbf{0.679}}} &\multirow{2}*{\textcolor{blue}{\textbf{0.373}}} &\multirow{2}*{0.664} &\multirow{2}*{0.323 }
            &\multirow{2}*{0.753 } &\multirow{2}*{\textcolor{blue}{\textbf{1.154}}} &\multirow{2}*{0.708 } &\multirow{2}*{0.850 }\\
        \specialrule{0em}{-0.5pt}{-2pt}
        {\textit{(Softmax)}}&&&&&&&&\\
        \specialrule{0em}{-0.5pt}{-1pt}
        \cline{1-9}
        Implicit attention
            &\multirow{2}*{0.642 } &\multirow{2}*{0.326 } &\multirow{2}*{0.707 } &\multirow{2}*{0.475 }
            &\multirow{2}*{0.685 } &\multirow{2}*{1.027 } &\multirow{2}*{0.726 } &\multirow{2}*{1.047 }\\
        \specialrule{0em}{-0.5pt}{-2pt}
        {\textit{(Sigmoid)}}&&&&&&&&\\
        \specialrule{0em}{-0.5pt}{-1pt}
        \cline{1-9}
        Explicit attention
            &\multirow{2}*{0.676 } &\multirow{2}*{0.359 } &\multirow{2}*{0.703 } &\multirow{2}*{0.457 }
            &\multirow{2}*{\textcolor{blue}{\textbf{0.754}}} &\multirow{2}*{1.150 } &\multirow{2}*{0.716 } &\multirow{2}*{\textcolor{blue}{\textbf{1.218}}}\\
        \specialrule{0em}{-0.5pt}{-2pt}
        {\textit{(Supervised)}}&&&&&&&&\\
        \specialrule{0em}{-0.5pt}{-1pt}
        \cline{1-9}
        Explicit attention
            &\multirow{2}*{\textcolor{red}{\textbf{0.918}} } &\multirow{2}*{\textcolor{red}{\textbf{2.225}} } &\multirow{2}*{\textcolor{red}{\textbf{0.884}} } &\multirow{2}*{\textcolor{red}{\textbf{1.486}} }
            &\multirow{2}*{\textcolor{red}{\textbf{0.909}} } &\multirow{2}*{\textcolor{red}{\textbf{2.547}} } &\multirow{2}*{\textcolor{red}{\textbf{0.864}} } &\multirow{2}*{\textcolor{red}{\textbf{1.753}} }\\
        \specialrule{0em}{-0.5pt}{-2pt}
        {\textit{(Human)}}&&&&&&&&\\
        \specialrule{0em}{-0.5pt}{-0pt}
        \hline
        \end{tabular}
    }
    \caption{The correlation between human and neural attentions on action recognition datasets (see \S\ref{sec:task2}).  
    \label{table:ActRec_Attention}}

\end{table}

\noindent\textbf{Correlation b/w positive and negative cases.} In Table~\ref{table:ActRec_Attention_statistics}, we note a different trend compared to Task1 (Table~\ref{table:SalObj_Attention_statistics}). Specifically, the correlation between attention maps for positive cases is generally lower compared to negative cases. This demonstrates that cues that assist in correct video action recognition are not concentrated in the same spatial locations. In contrast, the negative attentions are relatively more spatially dispersed and therefore have higher correlations among themselves.

\subsection{Task3: Fine-Grained Image Classification}\label{sec:task3}
Fine-grained classification aims at distinguishing the subtle differences among closely related classes.
Since humans attend to local details for fine-grained tasks, we are interested in how artificial and human attention compare on this complex task. 

\begin{figure}[t]
  \centering
    \includegraphics[width=0.6\linewidth]{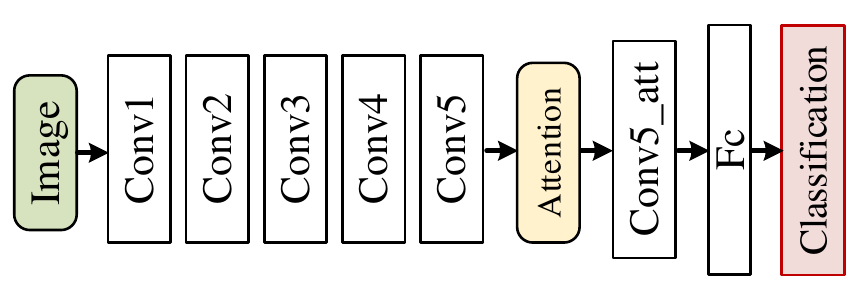}
  \caption{Illustration of network architectures for fine-grained image classification. 
  See \S\ref{sec:task3} for more details.
  \label{fig:FGC_network}}
\end{figure}

\subsubsection{Network Architectures}

We fine-tune AlexNet~\cite{krizhevsky2012imagenet}, VGGNet~\cite{simonyan2014very} and ResNet~\cite{he2016deep} for fine-grained image classification, where the number of the last fully-connected layer is adapted to the number of classes in the fine-grained dataset. For VGGNet, we add two dropout layers after the $1st$ and $2nd$ fully connected layers, respectively, to avoid overfitting.

The attention module is inserted before the last pooling layer, and is built as: \textit{Conv}($3\!\times\!3, \lfloor\frac{C}{2}\rfloor$) $\rightarrow$ \textit{ReLU} $\rightarrow$ \textit{BN} $\rightarrow$ \textit{Conv}($1\!\times\!1, 1$) $\rightarrow$ \textit{Sigmoid}, where $C$ is the channel number of the input feature. For \textit{Implicit Attention} (\textit{Softmax}) baseline, the single channel attention map is further normalized over all the spatial coordinates by applying the softmax.
For \textit{Explicit Attention} (\textit{Supervised}) baseline, the learned attention map is supervised by the ground-truth gaze maps. For \textit{Explicit Attention} (\textit{Human}), the artificial attention map is directly replaced by the ground-truth human attention map. The network architecture is shown in Fig.~\ref{fig:FGC_network}.

\subsubsection{Implementation Details}

\noindent\textbf{Datasets.}
In this task, we use \textit{CUB-VWSW}~\cite{karessli2017gaze} which contains $1,882$ images of $60$ classes selected from \textit{Caltech-UCSD Birds 200-2010}~\cite{WelinderEtal2010}, and gaze data collected from 5 participants during fine-grained learning between two images from different classes, followed by a classification process for determining which of the two classes a new instance belongs to.

\noindent\textbf{Training details.}
We follow the recommended train/text splitting\footnote{http://www.vision.caltech.edu/visipedia/CUB-200.html}, and use the subset of images that belongs to the $60$ classes accompanied with gaze data. We cropped the main object in each image using ground-truth bounding box annotations, and performed corresponding preprocessing, \textit{e.g.} resizing to $224\!\times\!224$ for VGG16 and ResNet50, and $227\!\times\!227$ for AlexNet, respectively, and subtracting image mean, \textit{.etc}. We implemented all the baselines using \textit{Keras}, and choose \textit{Adam}~\cite{kinga2015method} as the optimizer, where the initial learning rate is set to $5\times 10^{-6}, 10^{-5}$ and $10^{-4}$ for AlexNet, VGG16 and ResNet, respectively. The networks are initialized with weights trained on ImageNet~\cite{imagenet_cvpr09}. All the layers before fully connected layers are frozen during fine-tuning except the attention module.

We use \textit{foolbox}~\cite{rauber2017foolbox}\footnote{https://github.com/bethgelab/foolbox} to generate Fast Gradient Sign Method (FGSM)~\cite{goodfellow2014explaining} perturbations on the input images. 

\noindent\textbf{Evaluation Metrics.}

\textbf{\textit{Accuracy.}} In fine grained image classification, we calculate the accuracy as:
\begin{equation}\label{equation:fg_acc}
  Accuracy= \frac{\# correctly\;classified\;image}{\# total\;testing\;image}.
\end{equation}
In \textit{CUB-VWSW}, the $\# total\;testing\;image$ equals to $982$.

\textbf{\textit{Fooling rate.}} To measure the robustness of a model against the adversarial attacks, we calculate the fooling rate for each baseline, which indicates the percentage of the predicted labels that changes after the images are perturbed:
\begin{equation}\label{equation:foolingrate}
  Fooling\;rate= \frac{\# changed\;label}{\# perturbed\;image}.
\end{equation}
A model with lower fooling rate is more robust against the adversarial attacks.

\begin{table}[t]
    \centering
    \begin{threeparttable}
        \resizebox{0.405\textwidth}{!}{
        \setlength\tabcolsep{6pt}
        \renewcommand\arraystretch{1.}
        \begin{tabular}{c||c|c|c}
        \rowcolor{mygray}
        \hline\thickhline
         &\multicolumn{3}{c}{CUB-VWSW~\cite{karessli2017gaze}} \\
         \cline{2-4}
        \rowcolor{mygray}
             &AlexNet &~VGG~ & ResNet\\
        \cline{2-4}
        \rowcolor{mygray}
        \multirow{-3}*{Attention}&Accuracy~$\uparrow$&Accuracy~$\uparrow$& Accuracy~$\uparrow$\\
        \hline
        \hline
        Implicit attention
            &\multirow{2}*{0.223} &\multirow{2}*{0.409}  &\multirow{2}*{0.534}\\
            \specialrule{0em}{-0.5pt}{-2pt}
        {\textit{(Activation)}}&&&\\
        \specialrule{0em}{-0.5pt}{-1pt}
        \hline
         Implicit attention
            &\multirow{2}*{0.244}  &\multirow{2}*{0.426}  &\multirow{2}*{0.548} \\
            \specialrule{0em}{-0.5pt}{-2pt}
        {\textit{(Softmax)}}&&&\\
        \specialrule{0em}{-0.5pt}{-1pt}
        \hline
        Implicit attention
            &\multirow{2}*{0.235}  &\multirow{2}*{0.423}  &\multirow{2}*{0.550} \\
            \specialrule{0em}{-0.5pt}{-2pt}
        {\textit{(Sigmoid)}}&&&\\
        \specialrule{0em}{-0.5pt}{-1pt}
        \hline
        Explicit attention
            &\multirow{2}*{\textcolor{blue}{\textbf{0.247}}} &\multirow{2}*{\textcolor{blue}{\textbf{0.430}}} &\multirow{2}*{\textcolor{blue}{\textbf{0.553}}} \\
            \specialrule{0em}{-0.5pt}{-2pt}
        {\textit{(Supervised)}}&&&\\
        \specialrule{0em}{-0.5pt}{-1pt}
        \hline
        Explicit attention
            &\multirow{2}*{\textcolor{red}{\textbf{0.256}}} &\multirow{2}*{\textcolor{red}{\textbf{0.448}}} &\multirow{2}*{\textcolor{red}{\textbf{0.556}}} \\
        \specialrule{0em}{-0.5pt}{-2pt}
        {\textit{(Human)}}&&&\\
        \hline
        \end{tabular}
    }

    \caption{Quantitative results of fine-grained image classification baselines. 
    See \S\ref{sec:task3} for details.
    \label{table:FGClas}}

    \end{threeparttable}
\end{table}

\subsubsection{Analyses}
\noindent\textbf{How useful is attention?} 
Similar to Task1 case (\S\ref{sec:task1}), we evaluate fine-grained image classification with various backbones and attention baselines (results shown in Table~\ref{table:FGClas}). As humans attend to subtle differences to recognize closely related species, explicitly using human gaze maps proves best for all three backbones. The second best performance is achieved by using human attention as a supervisory signal for neural attention. Among implicit attention methods, sigmoid and softmax variants achieve somewhat similar performances. Overall, automatically learned neural attention performs lower compared to human attention.

\begin{table}[t]
    \centering
    \begin{threeparttable}
        \resizebox{0.482\textwidth}{!}{
        \setlength\tabcolsep{5.pt}
        \renewcommand\arraystretch{1.}
\begin{tabular}{c||c c| c c |c c}
        \hline\thickhline
        \rowcolor{mygray}
        &\multicolumn{6}{c}{CUB-VWSW~\cite{karessli2017gaze}} \\
        \cline{2-7}
        \rowcolor{mygray}
        &\multicolumn{2}{c|}{AlexNet} & \multicolumn{2}{c|}{VGGNet} & \multicolumn{2}{c}{ResNet}\\
        \cline{2-7}
        \rowcolor{mygray}
        \multirow{-3}*{Attention} &\!s-AUC~$\uparrow$\! &IG~$\uparrow$ &\!s-AUC~$\uparrow$\! &IG~$\uparrow$ &\!s-AUC~$\uparrow$\! &IG~$\uparrow$ \\
        \hline
        \hline
        Implicit attention
            &\multirow{2}*{0.669}&\multirow{2}*{0.250}
            &\multirow{2}*{\textcolor{blue}{\textbf{0.692}}} &\multirow{2}*{0.590}
            &\multirow{2}*{\textcolor{blue}{\textbf{0.692}}} &\multirow{2}*{0.410}\\
            \specialrule{0em}{-0.5pt}{-2pt}
        {\textit{(Activation)}}&&&&&&\\
        \specialrule{0em}{-0.5pt}{-1pt}
      \hline
        Implicit attention
             &\multirow{2}*{0.676} &\multirow{2}*{0.389}
             &\multirow{2}*{0.683} &\multirow{2}*{0.615}
             &\multirow{2}*{0.674} &\multirow{2}*{0.314} \\
             \specialrule{0em}{-0.5pt}{-2pt}
        {\textit{(Softmax)}}&&&&&&\\
        \specialrule{0em}{-0.5pt}{-1pt}
        \hline
        Implicit attention
             &\multirow{2}*{0.671}&\multirow{2}*{0.386}
             &\multirow{2}*{0.687} &\multirow{2}*{0.612}
             &\multirow{2}*{0.666} &\multirow{2}*{0.157} \\
             \specialrule{0em}{-0.5pt}{-2pt}
        {\textit{(Sigmoid)}}&&&&&&\\
        \specialrule{0em}{-0.5pt}{-1pt}
        \hline
        Explicit attention
            &\multirow{2}*{\textcolor{blue}{\textbf{0.678}}} &\multirow{2}*{\textcolor{blue}{\textbf{0.456}}} &\multirow{2}*{0.674} &\multirow{2}*{\textcolor{blue}{\textbf{0.703}}} &\multirow{2}*{0.654} &\multirow{2}*{\textcolor{blue}{\textbf{0.690}}} \\
            \specialrule{0em}{-0.5pt}{-2pt}
        {\textit{(Supervised)}}&&&&&&\\
        \specialrule{0em}{-0.5pt}{-1pt}
        \hline
        Explicit attention
            &\multirow{2}*{\textcolor{red}{\textbf{0.866}}} &\multirow{2}*{\textcolor{red}{\textbf{2.115}}} &\multirow{2}*{\textcolor{red}{\textbf{0.878}}} &\multirow{2}*{\textcolor{red}{\textbf{2.206}}} &\multirow{2}*{\textcolor{red}{\textbf{0.808}}} &\multirow{2}*{\textcolor{red}{\textbf{1.661}}} \\
            \specialrule{0em}{-0.5pt}{-2pt}
      {\textit{(Human)}}&&&&&&\\
        \hline
        \end{tabular}

    }
    \end{threeparttable}

    \caption{The correlation between human and neural attentions on fine-grained image classification dataset (\S\ref{sec:task3}). 
    }
    \label{table:FGClas_Attention}
\end{table}

\begin{figure*}[!htp]
\begin{minipage}{\textwidth}
 \begin{minipage}[t]{0.561\textwidth}
  \centering
    \begin{threeparttable}
        \resizebox{1\textwidth}{!}{
        \setlength\tabcolsep{2.pt}
        \renewcommand\arraystretch{1.}
        \begin{tabular}{c||c c |c c |c c |c c |c c |c c}
        \hline\thickhline
        \rowcolor{mygray}
        &\multicolumn{12}{c}{CUB-VWSW~\cite{karessli2017gaze}} \\
        \cline{2-13}
        \rowcolor{mygray}
        &\multicolumn{4}{c|}{AlexNet} & \multicolumn{4}{c|}{VGGNet} & \multicolumn{4}{c}{ResNet}\\
        \cline{2-13}
        \rowcolor{mygray}
         &\multicolumn{2}{c|}{\!s-AUC~$\uparrow$\!} &\multicolumn{2}{c|}{\!IG~$\uparrow$}
        &\multicolumn{2}{c|}{\!s-AUC~$\uparrow$\!} &\multicolumn{2}{c|}{IG~$\uparrow$}
        &\multicolumn{2}{c|}{\!s-AUC~$\uparrow$\!} &\multicolumn{2}{c}{IG~$\uparrow$} \\
        \cline{2-13}
        \rowcolor{mygray}
        \multirow{-4}*{Attention}&pos. &neg. &pos. &neg. &pos. &neg. &pos. &neg. &pos. &neg. &pos. &neg. \\
        \hline
        \hline
        Implicit attention
            &\multirow{2}*{0.661}&\multirow{2}*{0.671} &\multirow{2}*{0.190}&\multirow{2}*{0.267}
            &\multirow{2}*{0.702} &\multirow{2}*{0.685} &\multirow{2}*{0.613}&\multirow{2}*{0.574}
            &\multirow{2}*{0.697} &\multirow{2}*{0.687} &\multirow{2}*{0.419}&\multirow{2}*{0.399} \\
            \specialrule{0em}{-0.5pt}{-2pt}
        {\textit{(Activation)}}&&&&&&&&&&&&\\
        \specialrule{0em}{-0.5pt}{-1pt}
       \hline
        Implicit attention
             &\multirow{2}*{0.665} &\multirow{2}*{0.680} &\multirow{2}*{0.359}&\multirow{2}*{0.399}
             &\multirow{2}*{0.670} &\multirow{2}*{0.694} &\multirow{2}*{0.629}&\multirow{2}*{0.604}
             &\multirow{2}*{0.667} &\multirow{2}*{0.682} &\multirow{2}*{0.279}&\multirow{2}*{0.357} \\
             \specialrule{0em}{-0.5pt}{-2pt}
        {\textit{(Softmax)}}&&&&&&&&&&&&\\
        \specialrule{0em}{-0.5pt}{-1pt}
        \hline
        Implicit attention
             &\multirow{2}*{0.662} &\multirow{2}*{0.674} &\multirow{2}*{0.287}&\multirow{2}*{0.416}
             &\multirow{2}*{0.679} &\multirow{2}*{0.694} &\multirow{2}*{0.673}&\multirow{2}*{0.567}
             &\multirow{2}*{0.667} &\multirow{2}*{0.665} &\multirow{2}*{0.155}&\multirow{2}*{0.159} \\
             \specialrule{0em}{-0.5pt}{-2pt}
        {\textit{(Sigmoid)}}&&&&&&&&&&&&\\
        \specialrule{0em}{-0.5pt}{-1pt}
        \hline
        Explicit attention
            &\multirow{2}*{0.665} &\multirow{2}*{0.682}
            &\multirow{2}*{0.444} &\multirow{2}*{0.460}
            &\multirow{2}*{0.665} &\multirow{2}*{0.681}
            &\multirow{2}*{0.682} &\multirow{2}*{0.719}
            &\multirow{2}*{0.654} &\multirow{2}*{0.655}
            &\multirow{2}*{0.684} &\multirow{2}*{0.697} \\
            \specialrule{0em}{-0.5pt}{-2pt}
        {\textit{(Supervised)}}&&&&&&&&&&&&\\
        \specialrule{0em}{-0.5pt}{-1pt}
        \hline
        Explicit attention
            &\multirow{2}*{0.848} &\multirow{2}*{0.873}
            &\multirow{2}*{2.045} &\multirow{2}*{2.140}
            &\multirow{2}*{0.873} &\multirow{2}*{0.882}
            &\multirow{2}*{2.209} &\multirow{2}*{2.204}
            &\multirow{2}*{0.811} &\multirow{2}*{0.805}
            &\multirow{2}*{1.675} &\multirow{2}*{1.645} \\
            \specialrule{0em}{-0.5pt}{-2pt}
	    {\textit{(Human)}}&&&&&&&&&&&&\\
        \hline
        \end{tabular}
    }
    \makeatletter\def\@captype{table}\makeatother\caption{The correlation between positive and negative attention maps for fine-grained image classification dataset 
     (see \S\ref{sec:task3}).
    \label{table:FGClas_Attention_statistics}}
    \end{threeparttable}
  \end{minipage}
\begin{minipage}[t]{0.439\textwidth}
   \centering
         \begin{threeparttable}
        \resizebox{1\textwidth}{!}{
        \setlength\tabcolsep{4.5pt}
        \renewcommand\arraystretch{1.}
\begin{tabular}{c||c|c|c}
        \rowcolor{mygray}
        \hline\thickhline
         &\multicolumn{3}{c}{CUB-VWSW~\cite{karessli2017gaze}} \\
         \cline{2-4}
        \rowcolor{mygray}
             &AlexNet &~VGG~ & ResNet\\
        \cline{2-4}
        \rowcolor{mygray}
        \multirow{-3}*{Attention}&Fooling~rate~$\downarrow$&Fooling~rate~$\downarrow$& Fooling~rate~$\downarrow$\\
        \hline
        \hline
        Implicit attention
            &\multirow{2}*{0.5251} &\multirow{2}*{0.6368} &\multirow{2}*{0.7958}  \\
        \specialrule{0em}{-0.5pt}{-2pt}
        {\textit{(Activation)}}&&&\\
        \specialrule{0em}{-0.5pt}{-1pt}
        \hline
        Implicit attention
            &\multirow{2}*{0.5208} &\multirow{2}*{0.6148} &\multirow{2}*{0.8082}  \\
        \specialrule{0em}{-0.5pt}{-2pt}
        {\textit{(Softmax)}}&&&\\
        \specialrule{0em}{-0.5pt}{-1pt}
        \hline
        Implicit attention
            &\multirow{2}*{0.5368} &\multirow{2}*{\textcolor{blue}{\textbf{0.5880}}} &\multirow{2}*{\textcolor{red}{\textbf{0.7611}}}\\
        \specialrule{0em}{-0.5pt}{-2pt}
        {\textit{(Sigmoid)}}&&&\\
        \specialrule{0em}{-0.5pt}{-1pt}
        \hline
        Explicit attention
            &\multirow{2}*{\textcolor{blue}{\textbf{0.4876}}} &\multirow{2}*{0.6019} &\multirow{2}*{0.7878} \\
        \specialrule{0em}{-0.5pt}{-2pt}
        {\textit{(Supervised)}}&&&\\
        \specialrule{0em}{-0.5pt}{-1pt}
        \hline
        Explicit attention
             &\multirow{2}*{\textcolor{red}{\textbf{0.4861}}}  &\multirow{2}*{\textcolor{red}{\textbf{0.5795}}} &\multirow{2}*{\textcolor{blue}{\textbf{0.7711}}} \\

    \specialrule{0em}{-0.5pt}{-2pt}
    {\textit{(Human)}}&&&\\
            \specialrule{0em}{-0.5pt}{-0pt}
        \hline
        \end{tabular}
    }
    \makeatletter\def\@captype{table}\makeatother\caption{Fooling rates of fine-grained image classification baselines under FGSM attack. 
    \label{table:FGClas_fooling_rate}}

    \end{threeparttable}
   \end{minipage}

  \end{minipage}
\end{figure*}

\begin{figure*}[!ht] 
	\centering
    \includegraphics[width=0.9\linewidth]{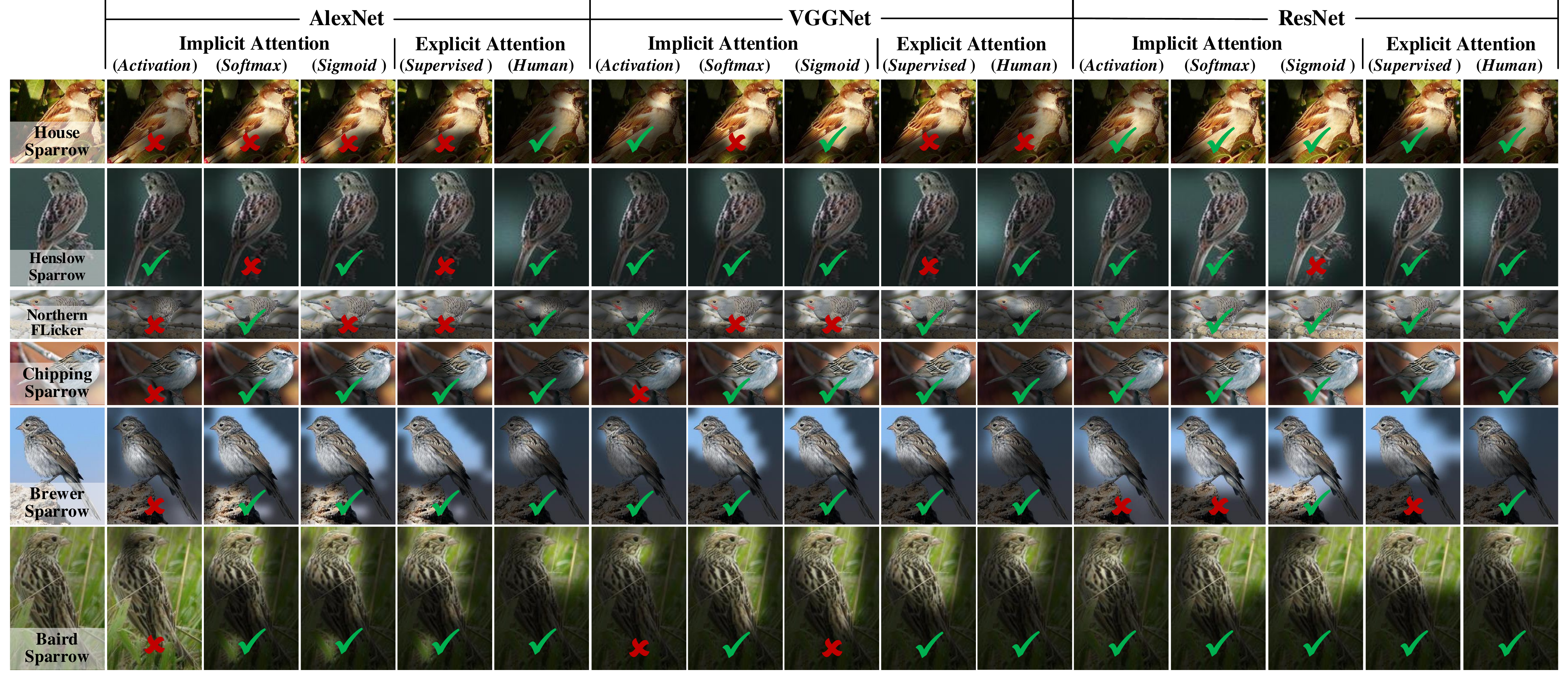}

	\caption{Qualitative Results of fine-grained classification baselines.  See \S\ref{sec:task3} for details.
	}
    \label{fig:FGC}
\end{figure*}

\noindent\textbf{Correlation b/w human and artificial attention.}
Similar to previous tasks, we use s-AUC~\cite{borji2013analysis} and IG~\cite{kummerer2015information} to compare human and artificial attention (see Table~\ref{table:FGClas_Attention}). From our results, we conclude that explicit human attention provides the best performance for fine-grained image classification tasks. The closest to direct human attention is the artificial attention learned using supervised human attention maps. Automatically learned artificial attention with softmax performs best amongst other artificial attention mechanisms, but still performs significantly lower than the explicit attention mechanisms.

A qualitative comparison of different attention maps is shown in Fig.~\ref{fig:FGC}. As can be observed:

\noindent{\textbullet}~It would be more likely to reach a correct classification when the attention maps highlight the `important' region denoted by the top-down human attention. When looking at the failure cases, we notice that most of them failed to look at the crucial part of the bird (most often, the area near the head/beak).

\noindent{\textbullet}~Sometimes, even when the attention map looks correct, the recognition still fails. This is because the representation ability of the backbone also plays an important role for fine-grained image classification. With visually similar attention maps, ResNet baselines would perform better compared with AlexNet and VGGNet baselines for most of the cases. 

\noindent\textbf{Correlation b/w positive and negative cases:} Here, we are interested in studying how the attention for positive and negative cases compare with each other. We note two major trends from our results shown in Table~\ref{table:FGClas_Attention_statistics}. \textbf{(1)} Generally, one would assume that the attention for positive cases is more likely to correlate well, while the attention for negative cases may be mismatched from one case to another. In fact, we note an opposite trend. Our overall results show that, in most cases, the attention for negative cases is better correlated. This is because  discriminative information appear in different spatial regions even for similar examples. As a result, the positive attentions are more concentrated and differ from one example to another. In comparison, the attention for negative cases are more dispersed and do not differ much from one example to another. \textbf{(2)} The correlation strength increases from implicit attention to explicit attention mechanisms. This shows that human attentions tend to be more concentrated on specific regions (due to past experiences) while machine attention is more evenly spread (as it depends more on individual inputs).

\begin{figure*}[t]
	\centering
    \includegraphics[width=\linewidth]{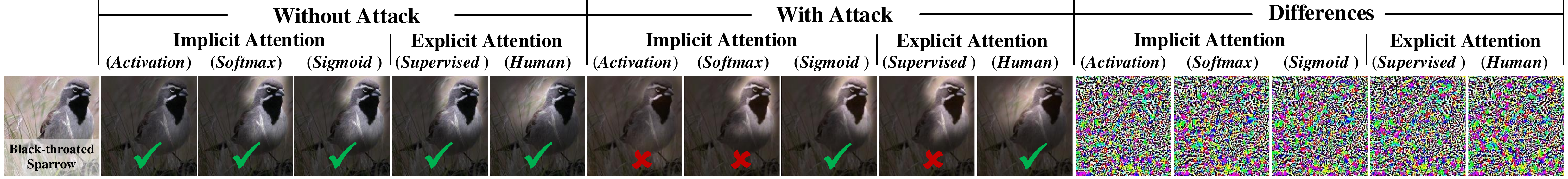}

	\caption{Visual results of VGG based fine-grained classification model \textit{w.} or \textit{w.o.} FGSM attack. See \S\ref{sec:task3} for details. 
	}
    \label{fig:FGC_attack}
\end{figure*}

\noindent\textbf{Robustness to adversarial attacks.}
Adversarial attacks add human imperceptible perturbations to the inputs that fool the neural networks. In this section, we are particularly interested in studying how human and artificial attention mechanisms perform against adversarial attacks. We consider FGSM~\cite{goodfellow2014explaining} to generate adversarial examples using original images. 
The fooling rate shown in Table~\ref{table:FGClas_fooling_rate} indicates that the introduction of human attention provides the highest robustness against adversarial attacks. This is intuitive, because the adversarial noise mainly aims to shift network attention to unrelated parts of an image and shifting the attention back to most significant image details helps in restoring the network confidences towards ground-truth classes. We also note that the fooling rate is greatly reduced for the networks that achieve low performance (\eg, ResNet \textit{v.s.} AlexNet) when explicit human attention is introduced. An example is shown in Fig.~\ref{fig:FGC_attack}.

\section{Further Insights}

\noindent
\textit{Q1: Are automatically learned attention maps close to top-down human attention? \\
A1:} Not really. The correlation analysis between human and artificial attentions shows that there exists a gap between artificial and human attention on all three studied tasks. For low-level attention-driven tasks such as salient object segmentation, human attention remarkably outperform artificial attention (Table~\ref{table:SalObj_Attention}). But it is case-by-case for high-level tasks. For closely attention-related tasks like fine-grained image classification, the better performance will be achieved when neural attention is closer to human attention (Table~\ref{table:FGClas_Attention}). While for other tasks like action recognition, this gap is relatively narrow (Table~\ref{table:ActRec_Attention}). This reveals that artificial attention mechanisms are not guaranteed to be close to top-down human attention.

\noindent
\textit{Q2: Can human attention serve as a relatively reliable and meaningful `groundtruth' for certain tasks? \\ 
A2:} It depends on the property of the task. Our evaluations show that human attention leads to better performance for saliency detection (Table~\ref{table:SalObj}) and fine-grained classification tasks (Table~\ref{table:FGClas}) compared with all other attention baselines, while does not show superiority in video action recognition (Table~\ref{table:ActReg}). We also observe that the explicit supervised attention (that uses human attention to learn artificial attention) performs best among all the artificial attention mechanisms in Task1 (Table~\ref{table:SalObj}) and Task3 (Table~\ref{table:FGClas}), which shows that a human attention consistent approach is generally helpful for the attention-driven or closely attention-related tasks. 

\noindent
\textit{Q3: Can we conclude that attention for correct predictions is concentrated on certain image regions?\\
A3:} We calculate the correlation of attention maps between top correct predictions and highly incorrect predictions for the three studied tasks. Interestingly, we note that for tasks that directly align with human perception (\eg, saliency detection), the correlation between correct prediction is higher (see Table~\ref{table:SalObj_Attention_statistics}). However for other tasks, the trend shows that useful cues are generally spread out and there are no fix spatial locations that are always helpful (see Tabs.~\ref{table:ActRec_Attention_statistics} and \ref{table:FGClas_Attention_statistics}).

\noindent
\textit{Q4: How do attention maps computed from different network types and depths vary?\\
A4:} Generally, the network type with a more effective representative ability will derive more powerful attention maps. For example, on both saliency segmentation (Table~\ref{table:SalObj}) and fine-grained classification (Table~\ref{table:FGClas}), ResNet outperforms other two backbones. Among VGGNet and AlexNet models, VGGNet performs better. When human attention is applied on more effective architectures like ResNet, it also helps achieve a more significant boost compared to VGGNet and AlexNet. 
The attention maps from different layers also vary in their properties. 
Take action recognition for example, early fusion of implicit attention mechanisms generally performs lower compared to late fusion. However, for explicit attention mechanisms, the trend is the opposite \ie, late fusion performs worse compared to early fusion (see Table~\ref{table:ActReg}).

\noindent
\textit{Q5: What can help model become closer to human attention?\\
A5:} Our experiments show that except the network depth and direct supervision from human attention, the choice of activation function to process artificial attention maps also helps in generating better attention maps. E.g., the sigmoid activation generally performs better for video action recognition (Table~\ref{table:ActReg}) while the softmax activation is overall a better choice for fine-grained recognition task (Table~\ref{table:FGClas}). 

\noindent
\textit{Q6: Is attending to correct visual details helpful in avoiding adversarial attacks?\\
A6:} Yes, our experiment with FGSM attack on fine-grained image classification shows that networks with different attention mechanism present a different robustness against the adversarial attacks (see Table~\ref{table:FGClas_fooling_rate}). We found that if a model attends to important visual details, robustness against adversarial attacks can be significantly improved.

\noindent
\textit{Q7: Finally, what is the way forward?\\
A7:} Human attention shows its capability in bench-marking the meaniful `groundtruth' for low-level attention-driven tasks. However, things are more complicated for high-level vision tasks; the task complexity may not allow a proper alignment between the neural and human attentions. We attribute this to the fact that learning objectives defined in deep networks directly focus on minimizing the error rate and do not consider the intermediate attentions that shape those decisions. 
An important consideration for future deep network design of low-level attention-driven tasks is to explicitly force a better alignment between artificial and human attentions. For high-level vision task, such alignment would also benefit making the deep networks more transparent and explainable. 

\section{Conclusion}

We provide an in-depth analysis for human and artificial attention mechanisms in deep neural networks. We address some of the pressing questions such as: if neural attention maps correspond to human eye fixations; if human attention can be the right benchmark for neural attention; how attention changes with network types and depths; if attention helps avoid adversarial attacks. Our study is supported by thorough experiments on three important computer vision tasks, namely saliency object segmentation, video action recognition, and fine-grained categorization. Our experiments show that human attention is valuable for deep networks to achieve better performance and enhance robustness against perturbations, especially for attention-driven tasks. The design of artificial attention mechanisms that can closely mimic human attention is a challenging task, but certainly a worthwhile endeavour.

\section*{Acknowledgment}

The authors would like to thank the anonymous reviewers for the valuable comments. This work is supported in part by the Natural Science Foundation of Guangdong Province, China, under Grant 2019A1515010860, in part by the Fundamental Research Funds for the Central Universities under Grant D2190670, in part by the Science and Technology Project of Guangzhou City under Grant 201707010140.


%



%
%
%
%



\bibliographystyle{IEEEtran}

\begin{thebibliography}{10}
\providecommand{\url}[1]{#1}
\csname url@samestyle\endcsname
\providecommand{\newblock}{\relax}
\providecommand{\bibinfo}[2]{#2}
\providecommand{\BIBentrySTDinterwordspacing}{\spaceskip=0pt\relax}
\providecommand{\BIBentryALTinterwordstretchfactor}{4}
\providecommand{\BIBentryALTinterwordspacing}{\spaceskip=\fontdimen2\font plus
\BIBentryALTinterwordstretchfactor\fontdimen3\font minus
  \fontdimen4\font\relax}
\providecommand{\BIBforeignlanguage}[2]{{%
\expandafter\ifx\csname l@#1\endcsname\relax
\typeout{** WARNING: IEEEtran.bst: No hyphenation pattern has been}%
\typeout{** loaded for the language `#1'. Using the pattern for}%
\typeout{** the default language instead.}%
\else
\language=\csname l@#1\endcsname
\fi
#2}}
\providecommand{\BIBdecl}{\relax}
\BIBdecl


\bibitem{koch2006much}
K.~Koch, J.~McLean, R.~Segev, M.~A. Freed, M.~J. Berry~II, V.~Balasubramanian,
  and P.~Sterling, ``How much the eye tells the brain,'' \emph{Current
  Biology}, vol.~16, no.~14, pp. 1428--1434, 2006.

\bibitem{eriksen1972temporal}
C.~W. Eriksen and J.~E. Hoffman, ``Temporal and spatial characteristics of
  selective encoding from visual displays,'' \emph{Perception \&
  Psychophysics}, vol.~12, no.~2, pp. 201--204, 1972.

\bibitem{treisman1980feature}
A.~M. Treisman and G.~Gelade, ``A feature-integration theory of attention,''
  \emph{Cognitive Psychology}, vol.~12, no.~1, pp. 97--136, 1980.

\bibitem{koch1987shifts}
C.~Koch and S.~Ullman, ``Shifts in selective visual attention: Towards the
  underlying neural circuitry,'' in \emph{Matters of Intelligence}, 1987, pp.
  115--141.

\bibitem{connor2004visual}
C.~E. Connor, H.~E. Egeth, and S.~Yantis, ``Visual attention: Bottom-up versus
  top-down,'' \emph{Current Biology}, vol.~14, no.~19, pp. 850--852, 2004.

\bibitem{itti1998model}
L.~Itti, C.~Koch, and E.~Niebur, ``A model of saliency-based visual attention
  for rapid scene analysis,'' \emph{IEEE Trans. Pattern Anal. Mach. Intell.}, 
  vol.~20, no.~11, pp. 1254--1259, 1998.

\bibitem{itti2001computational}
L.~Itti and C.~Koch, ``Computational modelling of visual attention,''
  \emph{Nature Reviews Neuroscience}, vol.~2, no.~3, p. 194, 2001.

\bibitem{borji2015salient}
A.~Borji, M.-M. Cheng, H.~Jiang, and J.~Li, ``Salient object detection: A
  benchmark,'' \emph{IEEE Trans. Image Process.}, vol.~24, no.~12, pp.
  5706--5722, 2015.

\bibitem{li2017benchmark}
J.~Li, C.~Xia, and X.~Chen, ``A benchmark dataset and saliency-guided stacked
  autoencoders for video-based salient object detection,'' \emph{IEEE Trans. 
  Image Process.}, vol.~27, no.~1, pp. 349--364, 2017.


\bibitem{liu2017distributed}
X.~Liu, Z.~Li, C.~Deng, and D.~Tao, ``Distributed adaptive binary quantization
  for fast nearest neighbor search,'' \emph{IEEE Trans. Image Process.},
  vol.~26, no.~11, pp. 5324--5336, 2017.

\bibitem{wang2018attentive}
W.~Wang, Y.~Xu, J.~Shen, and S.~Zhu, ``Attentive fashion grammar
  network for fashion landmark detection and clothing category classification,'' 
  in \emph{Proc. IEEE Conf. Comput. Vis. Pattern Recognit.}, 
  pp. 4271--4280, 2018.

\bibitem{gao2005discriminant}
D.~Gao and N.~Vasconcelos, ``Discriminant saliency for visual recognition from
  cluttered scenes,'' in \emph{Proc. Advances Neural Inf. Process. Syst.}, 
  2005, pp. 481--488.

\bibitem{sharma2012discriminative}
G.~Sharma, F.~Jurie, and C.~Schmid, ``Discriminative spatial saliency for image
  classification,'' in \emph{Proc. IEEE Conf. Comput. Vis. Pattern Recognit.}, 
  2012, pp. 3506--3513.

\bibitem{hadizadeh2012eye}
H.~Hadizadeh, M.~J. Enriquez, and I.~V. Bajic, ``Eye-tracking database for a
  set of standard video sequences,'' \emph{IEEE Trans. Image Process.},
  vol.~21, no.~2, pp. 898--903, 2012.


\bibitem{mathe2015actions}
S.~Mathe and C.~Sminchisescu, ``Actions in the eye: Dynamic gaze datasets and
  learnt saliency models for visual recognition,'' \emph{IEEE Trans. Pattern 
  Anal. Mach. Intell.}, vol.~37, no.~7, pp. 1408--1424, 2015.

\bibitem{wang2018saliency}
W.~Wang, J.~Shen, R.~Yang, and F.~Porikli, ``Saliency-aware video object
  segmentation,'' \emph{IEEE Trans. Pattern Anal. Mach. Intell.}, 
  vol.~40, no.~1, pp. 20--33, 2018.

\bibitem{tsotsos1995modeling}
J.~K. Tsotsos, S.~M. Culhane, W.~Y.K. Wai, Y.~Lai, N.~Davis, F.~ Nuflo,  
  ``Modeling visual attention via selective tuning,'' in \emph{Artificial 
  Intelligence}, vol.~78, no.~1-2, pp. 507--545, 1995

\bibitem{biparva2017stnet}
M.~Biparva, J.~Tsotsos, ``STNet: Selective Tuning of Convolutional Networks 
  for Object Localization'', in \emph{Proc. IEEE Int. Conf. Comput. Vis.}, 
  2017, pp. 2715--2723

\bibitem{zhang2018top} 
J.~Zhang, S.~A. Bargal, Z.~Lin, J.~Brandt, X.~Shen, S.~Sclaroff, Stan,
  ``Top-down neural attention by excitation backprop,'' in \emph{Int. 
  J. Comput. Vis.}, vol.~126, no.~10, pp. 1084--1102, 2018

\bibitem{bahdanau2014neural}
D.~Bahdanau, K.~Cho, and Y.~Bengio, ``Neural machine translation by jointly
  learning to align and translate,'' in \emph{Proc. Int. Conf. 
  Learn. Representations}, 2015.

\bibitem{rush2015neural}
A.~M. Rush, S.~Chopra, and J.~Weston, ``A neural attention model for
  abstractive sentence summarization,'' in \emph{Proceedings of Conference on
  Empirical Methods in Natural Language Processing}, 2015, pp. 379--389.

\bibitem{wang2016attention}
Y.~Wang, M.~Huang, L.~Zhao \emph{et~al.}, ``Attention-based {LSTM} for
  aspect-level sentiment classification,'' in \emph{Proceedings of Conference
  on Empirical Methods in Natural Language Processing}, 2016, pp. 606--615.

\bibitem{vaswani2017attention}
A.~Vaswani, N.~Shazeer, N.~Parmar, J.~Uszkoreit, L.~Jones, A.~N. Gomez,
  {\L}.~Kaiser, and I.~Polosukhin, ``Attention is all you need,'' in
  \emph{Proc. Advances Neural Inf. Process. Syst.}, 2017, pp.
  5998--6008.

\bibitem{xu2015show}
K.~Xu, J.~Ba, R.~Kiros, K.~Cho, A.~Courville, R.~Salakhudinov, R.~Zemel, and
  Y.~Bengio, ``Show, attend and tell: Neural image caption generation with
  visual attention,'' in \emph{Proc. Int. Conf. Learn. Representations},
  2015.

\bibitem{cao2015look}
C.~Cao, X.~Liu, Y.~Yang, Y.~Yu, J.~Wang, Z.~Wang, Y.~Huang, L.~Wang, C.~Huang,
  W.~Xu, D.~Ramanan, and T.~S. Huang, ``Look and think twice: Capturing
  top-down visual attention with feedback convolutional neural networks,'' in
  \emph{Proc. IEEE Conf. Comput. Vis. Pattern Recognit.}, 2015.

\bibitem{yang2016stacked}
Z.~Yang, X.~He, J.~Gao, L.~Deng, and A.~Smola, ``Stacked attention networks for
  image question answering,'' in \emph{Proc. IEEE Conf. Comput. 
  Vis. Pattern Recognit.}, 2016.

\bibitem{farazi2018reciprocal}
M.~M. Farazi and S.~Khan, ``Reciprocal attention fusion for visual question
  answering,'' \emph{British Machine Vision Conference}, 2018.

\bibitem{sharma2015action}
S.~Sharma, R.~Kiros, and R.~Salakhutdinov, ``Action recognition using visual
  attention,'' in \emph{Proc. Int. Conf. Learn. Representations Workshops.}, 2015.

\bibitem{girdhar2017attentional}
R.~Girdhar and D.~Ramanan, ``Attentional pooling for action recognition,'' in
  \emph{Proc. Advances Neural Inf. Process. Syst.}, 2017, pp. 34--45.

\bibitem{zhu2018fine}
C.~Zhu, X.~Tan, F.~Zhou, X.~Liu, K.~Yue, E.~Ding, and Y.~Ma, ``Fine-grained
  video categorization with redundancy reduction attention,'' in \emph{Proc. 
  Eur. Conf. Comput. Vis.}, 2018, pp. 136--152.

\bibitem{du2018interaction}
Y.~Du, C.~Yuan, B.~Li, L.~Zhao, Y.~Li, and W.~Hu, ``Interaction-aware
  spatio-temporal pyramid attention networks for action classification,'' in
  \emph{Proc. Eur. Conf. Comput. Vis.}, 2018, pp. 373--389.

\bibitem{Zhang_2018_CVPR}
X.~Zhang, T.~Wang, J.~Qi, H.~Lu, and G.~Wang, ``Progressive attention guided
  recurrent network for salient object detection,'' in \emph{Proc. 
  IEEE Conf. Comput. Vis. Pattern Recognit.}, 2018.

\bibitem{wang2018salient}
W.~Wang, J.~Shen, X.~Dong, and A.~Borji, ``Salient object detection driven by
  fixation prediction,'' in \emph{Proc. IEEE Conf. Comput. Vis. 
  Pattern Recognit.}, 2018, pp. 1711--1720.

\bibitem{Liu_2018_CVPR}
N.~Liu, J.~Han, and M.-H. Yang, ``Picanet: Learning pixel-wise contextual
  attention for saliency detection,'' in \emph{Proc. IEEE
  Conf. Comput. Vis. Pattern Recognit.}, 2018.

\bibitem{wang2019salient}
W.~Wang, S.~Zhao, J.~Shen, S.~C. Hoi, and A.~Borji, ``Salient object detection with 
  pyramid attention and salient edges,'' in \emph{Proc. IEEE 
  Conf. Comput. Vis. Pattern Recognit.}, pp. 1448--1457, 2019.

\bibitem{Chen_2018_ECCV}
S.~Chen, X.~Tan, B.~Wang, and X.~Hu, ``Reverse attention for salient object
  detection,'' in \emph{Proc. Eur. Conf. Comput. Vis.}, 2018.

\bibitem{Woo_2018_ECCV}
S.~Woo, J.~Park, J.-Y. Lee, and I.~So~Kweon, ``{CBAM}: Convolutional block
  attention module,'' in \emph{Proc. Eur. Conf. Comput. Vis.}, 2018.

\bibitem{wang2019learning}
W.~Wang, H.~Song, S.~Zhao, J.~Shen, S.~Zhao, S.~C. Hoi, and H.~Ling, ``Learning 
  unsupervised video object segmentation through visual attention,'' in 
  \emph{Proc. IEEE Conf. Comput. Vis. Pattern Recognit.}, 
  pp. 3064--3074, 2019.

\bibitem{das2016human}
A.~Das, H.~Agrawal, L.~Zitnick, D.~Parikh, and D.~Batra, ``Human attention in
  visual question answering: Do humans and deep networks look at the same
  regions?'' in \emph{Proceedings of Conference on Empirical Methods in Natural
  Language Processing}, 2016, pp. 932--937.

\bibitem{Lipton:2018:MMI:3236386.3241340}
Z.~C. Lipton, ``The mythos of model interpretability,'' \emph{Queue}, vol.~16,
  no.~3, pp. 30:31--30:57, 2018.

\bibitem{katsuki2014bottom}
F.~Katsuki and C.~Constantinidis, ``Bottom-up and top-down attention: Different
  processes and overlapping neural systems,'' \emph{The Neuroscientist},
  vol.~20, no.~5, pp. 509--521, 2014.

\bibitem{li2014secrets}
Y.~Li, X.~Hou, C.~Koch, J.~M. Rehg, and A.~L. Yuille, ``The secrets of salient
  object segmentation,'' in \emph{Proc. IEEE Conf. Comput. Vis. 
  Pattern Recognit.}, 2014, pp. 280--287.

\bibitem{yang2013saliency}
C.~Yang, L.~Zhang, H.~Lu, X.~Ruan, and M.-H. Yang, ``Saliency detection via
  graph-based manifold ranking,'' in \emph{Proc. IEEE Conf. 
  Comput. Vis. Pattern Recognit.}, 2013.

\bibitem{rodriguez2008action}
M.~D. Rodriguez, J.~Ahmed, and M.~Shah, ``Action mach a spatio-temporal maximum
  average correlation height filter for action recognition,'' in
  \emph{Proc. IEEE Conf. Comput. Vis. Pattern Recognit.}, 
  2008, pp. 1--8.

\bibitem{karessli2017gaze}
N.~Karessli, Z.~Akata, B.~Schiele, and A.~Bulling, ``Gaze embeddings for
  zero-shot image classification,'' in \emph{Proc. IEEE Conf. 
  Comput. Vis. Pattern Recognit.}, 2017.

\bibitem{xu2018structured}
D.~Xu, W.~Wang, H.~Tang, H.~Liu, N.~Sebe, and E.~Ricci``Structured attention 
  guided convolutional neural fields for monocular depth estimation,'' 
  in \emph{Proc. IEEE Conf. Comput. Vis. Pattern Recognit.}, 2018, pp. 3917--3925.

\bibitem{xie2019image}
C.~Xie, S.~Liu, C.~Li, M.-M. Cheng, W.~Zuo, X.~Liu, S.~Wen, and E.~Ding, ``Image 
  Inpainting with Learnable Bidirectional Attention Maps'', in \emph{Proc. IEEE 
  Int. Conf. Comput. Vis.}, 2019, pp. 8858--8867.

\bibitem{zhai2019optical}
M.~Zhai, X.~Xiang, R.~Zhang, N.~Lv, and A.~El Saddik, ``Optical Flow Estimation 
  Using Dual Self-Attention Pyramid Networks,'' \emph{IEEE Trans. Circuits Syst. 
  Video Technol.}, 2019.

\bibitem{wang2019sodsurvey}
W.~Wang, Q.~Lai, H.~Fu, J.~Shen, H.~Ling, and R.~Yang, ``Salient object detection in 
  the deep learning era: An in-depth survey,’’ \emph{arXiv preprint arXiv:1904.09146}, 2019.

\bibitem{simonyan2014two}
K.~Simonyan and A.~Zisserman, ``Two-stream convolutional networks for action
  recognition in videos,'' in \emph{Proc. Advances Neural Inf. Process. Syst.}, 
  2014, pp. 568--576.

\bibitem{krizhevsky2012imagenet}
A.~Krizhevsky, I.~Sutskever, and G.~E. Hinton, ``Imagenet classification with
  deep convolutional neural networks,'' in \emph{Proc. Advances Neural Inf. 
  Process. Syst.}, 2012, pp. 1097--1105.

\bibitem{simonyan2014very}
K.~Simonyan and A.~Zisserman, ``Very deep convolutional networks for
  large-scale image recognition,'' in \emph{Proc. Int. Conf. Learn. 
  Representations}, 2015.

\bibitem{he2016deep}
K.~He, X.~Zhang, S.~Ren, and J.~Sun, ``Deep residual learning for image
  recognition,'' in \emph{Proc. IEEE Conf. Comput. Vis. 
  Pattern Recognit.}, 2016, pp. 770--778.

\bibitem{past25}
M.~Carrasco, ``Visual attention: The past 25 years,'' \emph{Vision Research},
  vol.~51, no.~13, pp. 1484--1525, 2011.

\bibitem{wang2017deep}
W.~Wang, and J.~Shen,  ``Deep visual attention prediction,'' \emph{IEEE Trans.  
  Image Process.}, vol.~27, no.~5, pp. 2368--2378, 2017.

\bibitem{wolfe1989guided}
J.~M. Wolfe, K.~R. Cave, and S.~L. Franzel, ``Guided search: An alternative to
  the feature integration model for visual search.'' \emph{Journal of
  Experimental Psychology: Human Perception and Performance}, vol.~15, no.~3,
  p. 419, 1989.

\bibitem{hwang2009model}
A.~D. Hwang, E.~C. Higgins, and M.~Pomplun, ``A model of top-down attentional
  control during visual search in complex scenes,'' \emph{Journal of Vision},
  vol.~9, no.~5, pp. 25--25, 2009.

\bibitem{pinto2013bottom}
Y.~Pinto, A.~R. van~der Leij, I.~G. Sligte, V.~A. Lamme, and H.~S. Scholte,
  ``Bottom-up and top-down attention are independent,'' \emph{Journal of
  Vision}, vol.~13, no.~3, pp. 16--16, 2013.

\bibitem{navalpakkam2006integrated}
V.~Navalpakkam and L.~Itti, ``An integrated model of top-down and bottom-up
  attention for optimizing detection speed,'' in \emph{Proc. IEEE
  Conf. Comput. Vis. Pattern Recognit.}, 2006, pp. 2049--2056.

\bibitem{judd2012benchmark}
T.~Judd, F.~Durand, and A.~Torralba, ``A benchmark of computational models of
  saliency to predict human fixations,'' \emph{MIT Technical Report}, 2012.

\bibitem{judd2009learning}
T.~Judd, K.~Ehinger, F.~Durand, and A.~Torralba, ``Learning to predict where
  humans look,'' in \emph{Proc. IEEE Conf. Comput. Vis. 
  Pattern Recognit.}, 2009, pp. 2106--2113.

\bibitem{bruce2006saliency}
N.~Bruce and J.~Tsotsos, ``Saliency based on information maximization,'' in
  \emph{Proc. Advances Neural Inf. Process. Syst.}, 2006, pp. 155--162.

\bibitem{jiang2015salicon}
M.~Jiang, S.~Huang, J.~Duan, and Q.~Zhao, ``{SALICON}: Saliency in context,''
  in \emph{Proc. IEEE Conf. Comput. Vis. Pattern Recognit.}, 
  2015, pp. 1072--1080.

\bibitem{mital2011clustering}
P.~K. Mital, T.~J. Smith, R.~L. Hill, and J.~M. Henderson, ``Clustering of gaze
  during dynamic scene viewing is predicted by motion,'' \emph{Cognitive
  Computation}, vol.~3, no.~1, pp. 5--24, 2011.

\bibitem{Wang_2018_CVPR}
W.~Wang, J.~Shen, F.~Guo, M.-M. Cheng, and A.~Borji, ``Revisiting video
  saliency: A large-scale benchmark and a new model,'' in \emph{Proc. IEEE 
  Conf. Comput. Vis. Pattern Recognit.}, 2018.

\bibitem{marszalek2009actions}
M.~Marszalek, I.~Laptev, and C.~Schmid, ``Actions in context,'' in
  \emph{Proc. IEEE Conf. Comput. Vis. Pattern Recognit.}, 
  2009, pp. 2929--2936.

\bibitem{simonyan2013deep}
K.~Simonyan, A.~Vedaldi, and A.~Zisserman, ``Deep inside convolutional
  networks: Visualising image classification models and saliency maps,''
  \emph{arXiv preprint arXiv:1312.6034}, 2013.

\bibitem{zhou2016learning}
B.~Zhou, A.~Khosla, A.~Lapedriza, A.~Oliva, and A.~Torralba, ``Learning deep
  features for discriminative localization,'' in \emph{Proc. IEEE
  Conf. Comput. Vis. Pattern Recognit.}, 2016, pp. 2921--2929.

\bibitem{zagoruyko2016paying}
S.~Zagoruyko and N.~Komodakis, ``Paying more attention to attention: Improving
  the performance of convolutional neural networks via attention transfer,'' in
  \emph{Proc. Int. Conf. Learn. Representations}, 2017.

\bibitem{mnih2014recurrent}
V.~Mnih, N.~Heess, A.~Graves, and K.~Kavukcuoglu, ``Recurrent models of visual
  attention,'' in \emph{Proc. Advances Neural Inf. Process. Syst.},
  2014, pp. 2204--2212.

\bibitem{song2017end}
S.~Song, C.~Lan, J.~Xing, W.~Zeng, and J.~Liu, ``An end-to-end spatio-temporal
  attention model for human action recognition from skeleton data,'' in
  \emph{AAAI Conference on Artificial Intelligence}, 2017.

\bibitem{wang2019zero}
W.~Wang, X.~Lu, J.~Shen, D.~Crandall, and L.~Shao, ``Zero-shot video object 
  segmentation via attentive graph neural networks,'' in \emph{Proc. IEEE Int. 
  Conf. Comput. Vis.}, pp. 9236--9245, 2019.

\bibitem{wangresidual}
F.~Wang, M.~Jiang, C.~Qian, S.~Yang, C.~Li, H.~Zhang, X.~Wang, and X.~Tang,
  ``Residual attention network for image classification,'' in \emph{Proc. IEEE 
  Conf. Comput. Vis. Pattern Recognit.}, 2017.

\bibitem{jetley2018learn}
S.~Jetley, N.~A. Lord, N.~Lee, and P.~H. Torr, ``Learn to pay attention,'' in
  \emph{Proc. Int. Conf. Learn. Representations}, 2018.


\bibitem{lu2016hierarchical}
J.~Lu, J.~Yang, D.~Batra, and D.~Parikh, ``Hierarchical question-image
  co-attention for visual question answering,'' in \emph{Proc. Advances 
  Neural Inf. Process. Syst.}, 2016, pp. 289--297.

\bibitem{bylinskii2016should}
Z.~Bylinskii, A.~Recasens, A.~Borji, A.~Oliva, A.~Torralba, and F.~Durand,
  ``Where should saliency models look next?'' in \emph{Proc. Eur. Conf. 
  Comput. Vis.}, 2016, pp. 809--824.

\bibitem{borji2013analysis}
A.~Borji, H.~R. Tavakoli, D.~N. Sihite, and L.~Itti, ``Analysis of scores,
  datasets, and models in visual saliency prediction,'' in \emph{Proc. IEEE 
  Conf. Comput. Vis. Pattern Recognit.}, 2013, pp. 921--928.

\bibitem{kummerer2015information}
M.~K{\"u}mmerer, T.~S. Wallis, and M.~Bethge, ``Information-theoretic model
  comparison unifies saliency metrics,'' \emph{Proceedings of the National
  Academy of Sciences}, vol. 112, no.~52, pp. 16\,054--16\,059, 2015.

\bibitem{Zhang_2018_ECCV}
P.~Zhang, J.~Xue, C.~Lan, W.~Zeng, Z.~Gao, and N.~Zheng, ``Adding attentiveness
  to the neurons in recurrent neural networks,'' in \emph{Proc. Eur. 
  Conf. Comput. Vis.}, 2018.

\bibitem{Bylinskiiwhat}
Z.~Bylinskii, T.~Judd, A.~Oliva, A.~Torralba, and F.~Durand, ``What do
  different evaluation metrics tell us about saliency models?'' \emph{IEEE
  Trans. Pattern Anal. Mach. Intell.}, 2018.

\bibitem{Everingham2010The}
M.~Everingham, L.~V. Gool, C.~K.~I. Williams, J.~Winn, and A.~Zisserman, ``The
  pascal visual object classes ({VOC}) challenge,'' \emph{Int. J. Comput. Vis.}, 
  vol.~88, no.~2, pp. 303--338, 2010.

\bibitem{imagenet_cvpr09}
J.~Deng, W.~Dong, R.~Socher, L.-J. Li, K.~Li, and L.~Fei-Fei, ``{ImageNet: A
  Large-Scale Hierarchical Image Database},'' in \emph{CVPR}, 2009, pp.
  248--255.

\bibitem{kinga2015method}
D.~Kinga and J.~B. Adam, ``A method for stochastic optimization,'' in
  \emph{ICLR}, vol.~5, 2015.

\bibitem{achanta2009frequency}
R.~Achanta, S.~Hemami, F.~Estrada, and S.~Susstrunk, ``Frequency-tuned salient
  region detection,'' in \emph{Proc. IEEE Conf. Comput. Vis. Pattern Recognit.}, 
  2009, pp. 1597--1604.

\bibitem{tran2015learning}
D.~Tran, L.~Bourdev, R.~Fergus, L.~Torresani, and M.~Paluri, ``Learning
  spatiotemporal features with 3d convolutional networks,'' in
  \emph{Proc. IEEE Conf. Comput. Vis. Pattern Recognit.}, 2015, pp. 4489--4497.

\bibitem{zach2007duality}
C.~Zach, T.~Pock, and H.~Bischof, ``A duality based approach for realtime tv-l
  1 optical flow,'' in \emph{Joint Pattern Recognition Symposium}, 2007, pp.
  214--223.

\bibitem{jia2014caffe}
Y.~Jia, E.~Shelhamer, J.~Donahue, S.~Karayev, J.~Long, R.~Girshick,
  S.~Guadarrama, and T.~Darrell, ``Caffe: Convolutional architecture for fast
  feature embedding,'' \emph{arXiv preprint arXiv:1408.5093}, 2014.

\bibitem{lecun1998gradient}
Y.~LeCun, L.~Bottou, Y.~Bengio, and P.~Haffner, ``Gradient-based learning
  applied to document recognition,'' \emph{Proceedings of the IEEE}, vol.~86,
  no.~11, pp. 2278--2324, 1998.

\bibitem{WelinderEtal2010}
P.~Welinder, S.~Branson, T.~Mita, C.~Wah, F.~Schroff, S.~Belongie, and
  P.~Perona, ``{Caltech-UCSD Birds 200},'' California Institute of Technology,
  Tech. Rep. CNS-TR-2010-001, 2010.

\bibitem{rauber2017foolbox}
\BIBentryALTinterwordspacing
J.~Rauber, W.~Brendel, and M.~Bethge, ``Foolbox: A python toolbox to benchmark
  the robustness of machine learning models,'' \emph{arXiv preprint
  arXiv:1707.04131}, 2017. [Online]. Available:
  \url{http://arxiv.org/abs/1707.04131}
\BIBentrySTDinterwordspacing

\bibitem{goodfellow2014explaining}
I.~J. Goodfellow, J.~Shlens, and C.~Szegedy, ``Explaining and harnessing
  adversarial examples,'' in \emph{Proc. Int. Conf. Learn. Representations}, 2015.

\end{thebibliography}
%


%
%

\vspace{-8mm}
\begin{IEEEbiographynophoto}
{Qiuxia Lai} received the B.E. and M.S. degrees in the School of Automation from Huazhong University of Science and Technology in 2013 and 2016, respectively. She is currently pursuing the Ph.D. degree with the Department of Computer Science and Engineering, The Chinese University of Hong Kong. Her research interests include image/video processing and deep learning.
\end{IEEEbiographynophoto}
\vspace{-8mm}
\begin{IEEEbiographynophoto}
{Salman Khan} is an Assistant Professor at the the Mohamed bin Zayed University of Artificial Intelligence, Abu Dhabi, UAE. Previously, he was a Research Scientist with Data61, Commonwealth Scientific and Industrial Research Organization (CSIRO), Australia from 2016 to 2018. He has also been an Honorary Lecturer with the College of Engineering and Computer Science at the Australian National University (ANU) since 2016.
His research interests include computer vision, pattern recognition, machine learning.
\end{IEEEbiographynophoto}
\vspace{-8mm}
\begin{IEEEbiographynophoto}
{Yongwei Nie} received the B.Sc. and Ph.D. degrees from Computer School, Wuhan University, in 2009 and 2015, respectively. He is currently an Associate Professor with the School of Computer Science and Engineering, South China University of Technology. His research interests include image and video editing, and computational photography.
\end{IEEEbiographynophoto}
\vspace{-8mm}
\begin{IEEEbiographynophoto}
{Hanqiu Sun} is the visiting professor at Shanghai Jiao-tong University and Zhejiang University, adjunct professor at University of ESTC. Her current research interests include virtual reality, interactive graphics, real-time hypermedia, virtual surgery, effective image/video synopsis, panoramic video displays, dynamics and touch-enhanced simulations. Sun has an MS in electrical engineering from University of British Columbia, and PhD in computer science from University of Alberta, Canada. 
\end{IEEEbiographynophoto}
\vspace{-8mm}
\begin{IEEEbiographynophoto}
{Jianbing Shen} (M'11-SM'12) is currently acting as the Lead Scientist at the Inception Institute of Artificial Intelligence, Abu Dhabi, UAE. He is also an adjunct Professor with the School of Computer Science, Beijing Institute of Technology.
He has published more than 100 journal and conference papers such as \textit{IEEE TIP}, \textit{IEEE TPAMI}, \textit{CVPR}, and \textit{ICCV}.
His research interests include computer vision and deep learning.
He is an Associate Editor of \textit{IEEE TIP}, \textit{IEEE TNNLS}, and other journals.
\end{IEEEbiographynophoto}
\vspace{-8mm}
\begin{IEEEbiographynophoto}
{Ling Shao} (M'09-SM'10) is the CEO and Chief Scientist of the Inception Institute of Artificial Intelligence, Abu Dhabi, United Arab Emirates. He is an Associate Editor of \textit{IEEE TIP}, \textit{IEEE TNNLS}, \textit{IEEE TCSVT}, and other journals. His research interests include Computer Vision, Machine Learning and Medical Imaging.
He is a Fellow of the IAPR, the IET and the BCS.
\end{IEEEbiographynophoto}
\vfill
\end{document}